\documentclass[10pt,twocolumn,letterpaper]{article}

\usepackage{iccv}
\usepackage{times}
\usepackage{epsfig}
\usepackage{graphicx}
\usepackage{amsmath}
\usepackage{amssymb}
\usepackage{subcaption}
\usepackage{booktabs}
\usepackage{algorithm}
\usepackage{algorithmic}
\usepackage{amsmath,amssymb}

\usepackage{graphicx}
\usepackage{textcomp}
\usepackage{xcolor}

\usepackage{mathtools}
\usepackage{float}
\usepackage{graphicx}
\usepackage{bbm}
\usepackage{multicol}
\usepackage{listings}
\usepackage{soul,color}
\usepackage{gensymb}
\usepackage{enumitem}
\usepackage{pbox}
\usepackage{multirow}

\usepackage{cuted}
\usepackage{etoolbox}

\usepackage[breaklinks=true,bookmarks=false]{hyperref}

\usepackage{amssymb}
\usepackage{pifont}
\newcommand{\cmark}{\ding{51}}%
\iccvfinalcopy 

\def\iccvPaperID{7416}

\ificcvfinal\fi
\begin{document}

\title{LOKI: Long Term and Key Intentions for Trajectory Prediction}
\author{
Harshayu Girase$^{1,2,}$\thanks{Co-first author. The authorship order is randomly determined}$^{~,\dagger}$~\quad Haiming Gang$^{1,*}$~\quad Srikanth Malla$^{1}$~\quad Jiachen Li$^{1,2,}$\thanks{Work done during the internship at Honda Research Institute}\\
Akira Kanehara$^{3}$~\quad Karttikeya Mangalam$^{2}$~\quad Chiho Choi$^{1}$\\
$^1$Honda Research Institute USA~\quad
$^{2}$University of California, Berkeley~\quad
$^{3}$Honda R\&D Co., Ltd.\\
{\tt\small \{harshayugirase, jiachen\_li, mangalam\}@berkeley.edu \quad akira\_kanehara@jp.honda} \\{\tt\small\{hgang, smalla, cchoi\}@honda-ri.com}
}

\maketitle
\thispagestyle{empty}

\begin{abstract}
Recent advances in trajectory prediction have shown that explicit reasoning about agents' intent is important to accurately forecast their motion. However, the current research activities are not directly applicable to intelligent and safety critical systems. This is mainly because very few public datasets are available, and they only consider pedestrian-specific intents for a short temporal horizon from a restricted egocentric view. To this end, we propose LOKI (LOng term and Key Intentions), a novel large-scale dataset that is designed to tackle joint trajectory and intention prediction for heterogeneous traffic agents (pedestrians and vehicles) in an autonomous driving setting. The LOKI dataset is created to discover several factors that may affect intention, including i) agent’s own will, ii) social interactions, iii) environmental constraints, and iv) contextual information. We also propose a model that jointly performs trajectory and intention prediction, showing that recurrently reasoning about intention can assist with trajectory prediction. We show our method outperforms state-of-the-art trajectory prediction methods by upto $27\%$ and also provide a baseline for frame-wise intention estimation. The dataset is available at \url{https://usa.honda-ri.com/loki}

\end{abstract}

\section{Introduction}

\begin{figure}[t]
\begin{center}
    \includegraphics[width=0.48\textwidth]{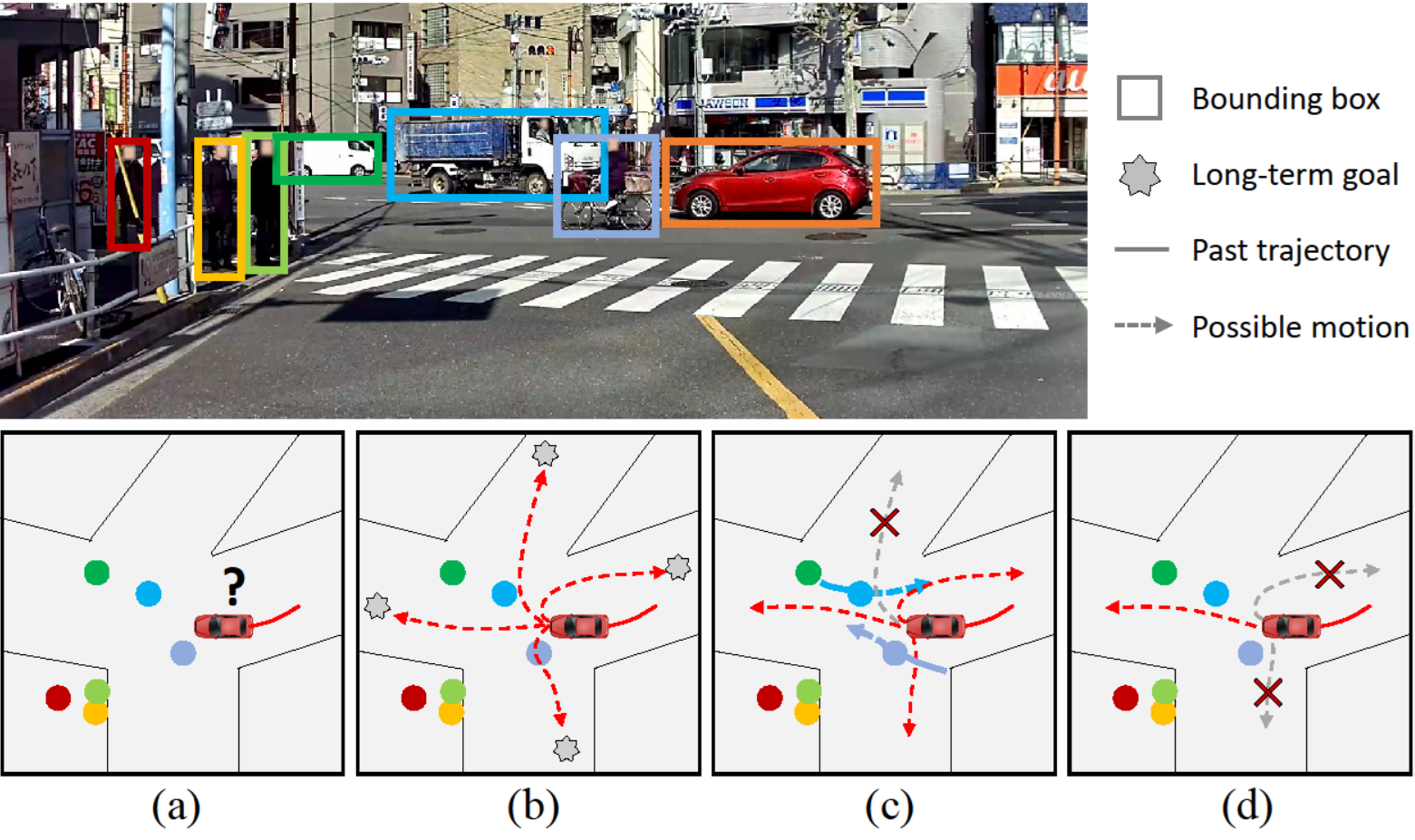}
    \caption{We show that reasoning about long-term goals and short-term intents plays a significant role in trajectory prediction. With a lack of comprehensive benchmarks for this purpose, we introduce a new dataset for intention and trajectory prediction. An example use case is illustrated in (a) where we predict the trajectory of the target vehicle. In (b), long-term goals are estimated from agent's own motion. Interactions in (c) and environmental constraints such as road topology and lane restrictions in (d) influence the agent's short-term intent and thus future trajectories.
    }
     \vspace{-0.6cm}
     \label{fig:main}
\end{center}
\end{figure}

Over the past few years, there has been extensive research into predicting future trajectories of dynamic agents in scenes, such as pedestrians and vehicles. This is an incredibly important and challenging task for safety critical applications such as autonomous vehicles or social robot navigation. While these methods have been significantly advanced over recent years, very few benchmarks specifically test if these models can accurately reason about key maneuvers such as sudden turns and lane changes of vehicles or pedestrians crossing the road. Traditional trajectory error metrics may not capture performance on frame-level maneuvers, which is critical for safe planning.

An intelligent trajectory prediction system should be able to understand and model dynamic human behaviors. The study of human behavior as goal-directed entities has a long and rich interdisciplinary history across the subfields of psychology \cite{carey1994domain}, neuroscience \cite{valentin2007determining} and computer vision \cite{mangalam2020goals}. The human decision making process is inherently hierarchical, consisting of several levels of reasoning and planning mechanisms that operate in tandem to achieving respective short and long term desires. Recent works have showed that explicitly reasoning about long-term goals \cite{mangalam2020goals, choi2019drogon, zhao2020tnt} and short-term intents \cite{liu2020spatiotemporal, rasouli2019pie, casas2018intentnet} can assist with trajectory prediction.

In this work, we propose to couple the tasks of heterogeneous (vehicles, pedestrians, etc.) multi-agent trajectory forecasting and intention prediction. We believe it is critical to explicitly reason about agents' long-term goals as well as their short-term intents. In our work, we define goals to be a final position an agent wants to reach for a given prediction horizon \cite{mangalam2020not, zhao2020tnt}, while intent refers to \textit{how} an agent accomplishes their goal \cite{rasouli2021pedestrian}. For example, consider a vehicle at an intersection. At the highest level, say they want to reach their ultimate goal of turning left to their final goal point, which in turn might be necessary for some higher-level end (such as going home). However, the exact motion of their trajectory is subject to many factors including i) agent's own will, ii) social interactions, iii) environmental constraints, iv) contextual cues. Thus, when reasoning about the agent's intent to turn left it is important to consider not only agent dynamics but also how intent is subject to change based on map topology or neighboring agents (see Figure~\ref{fig:main}). We believe this complex hierarchy of short-term intents and long-term goals is ubiquitous and in fact, crucial, for agent motion planning and hence by extension, for motion prediction. We propose an architecture that considers long-term goals similar to \cite{mangalam2020not, zhao2020tnt, mangalam2020goals, choi2019drogon} but adds a key component of frame-wise intention estimation which is used to condition the trajectory prediction module. By forcing the model to learn discrete short-term intents of agents, we observe improved performance by the prediction module. 

Equally rich \& successful is the contemporary history of the use of datasets for benchmarking progress in computer vision. Ushered by seminal works such as MNIST \cite{lecun1998mnist} and benchmarks such as ImageNet \cite{krizhevsky2012imagenet}, benchmarking progress and learning from data has played a key role in the success of modern deep learning. Currently, there exists no public datasets that allow for explicit frame-wise intention prediction for heterogeneous agents in highly complex environments. Although few datasets are designed to study pedestrian intents or actions \cite{rasouli2017they,  rasouli2019pie, liu2020spatiotemporal, malla2020titan} from egocentric view, it is an inherent limitation to extensive study of tasks for autonomous driving. Thus, we propose a joint trajectory and intention prediction dataset that contains RGB images with corresponding LiDAR point clouds with detailed, frame-wise labels for pedestrians \textit{and} vehicles. The LOKI dataset allows explicit modeling of agents' future intent and extensive benchmarking for both tasks. It also shows promising directions to jointly reason about intentions and trajectories while considering different external factors such as agents' predilection, social interactions and environmental factors. We show that by modeling short-term intent and long-term goals with explicit supervision via intention labels, better trajectory prediction accuracy can be achieved. In addition, predicting a specific intention at each frame adds a layer of abstraction to our model that improves understanding prediction decisions, an important step towards maintaining safety critical applications.

In conclusion, the contribution of our work is twofold. \textbf{First,} we propose the first publicly available heterogeneous dataset which contains frame-wise intention annotations and captures trajectories of up to 20 seconds containing both 2D and 3D labels with RGB and LiDAR inputs. \textbf{Second,} we illustrate the efficacy of separately reasoning about both long-term goals and short-term intents through ablation studies. Specifically, we highlight how the subtask of intention prediction improves prediction performance, and propose a model that outperforms state-of-the-art multimodal benchmarks by upto 27\%. We believe our highly flexible dataset will allow the trajectory prediction community to further explore topics within the  intention-based prediction space. In addition, the problem of intention estimation is an involved task in and of itself for which our work provides a strong baseline. 

\label{sec:introduction}
\section{Related Work}
 Over the past few years, there has been a rapid improvement in the field of trajectory prediction owing to the success of deep neural networks and larger publicly available datasets \cite{rudenko2020human, rasouli2019pie, liu2020spatiotemporal, rasouli2017they, chang2019argoverse, caesar2020nuscenes, kothari2020humantrajnet, sdd, 2019pedtrajpredsurvey}. There have been numerous subtopics of interest within the trajectory prediction community including compliant trajectory prediction, multi-modal trajectory prediction, and goal-oriented prediction \cite{li2020evolvegraph,sadeghian2019sophie, gupta2018social,ma2021continual,lisotto2019social, mangalam2020goals, mangalam2020not, kumar2020interaction, choi2019drogon, alahi2016social, deo2020trajectory, li2021rain,kosaraju2019social, choi2021shared,zhao2020tnt,girase2021physically}.

\subsection{Contextual Trajectory Prediction}
Earlier works in the field of trajectory prediction focused on unimodal trajectory prediction -- predicting a single future path for each agent. These works underscored the importance of social \cite{alahi2014socially, alahi2016social, helbing1995social} and scene compliance \cite{yagi2018future} when making predictions. Over the past few years, trajectory prediction studies have extended these ideas to multi-modal frameworks to account for multiple plausible futures each  agent can have. In SocialGAN, Gupta et al. \cite{gupta2018social} introduce a socially-aware multi-modal framework that uses generative adversarial networks to sample a varying number of future trajectories for each agent. Since then, there has been a major emphasis and many interesting approaches to with multimodal forecasting \cite{sadeghian2019sophie, lee2017desire, mangalam2020not, gupta2018social, li2019conditional, choi2019drogon, kosaraju2019social, salzmann2020trajectron++}.

\subsection{Goal-based Prediction}
When modeling vehicle and human trajectories, it is natural to formulate the problem as a goal-directed task. Because humans are not completely stochastic agents and have a predilection towards certain actions, very recent trajectory forecasting studies have shown the effectiveness of goal-conditioned predictions \cite{rhinehart2019precog, deo2020trajectory, mangalam2020not, mangalam2020goals, yao2021bitrap, rasouli2019pie, zhang2020map,choi2019drogon, zhao2020tnt, dendorfer2020goal}. Recently,   \cite{mangalam2020not} and \cite{zhao2020tnt} showed that considering agents' final goal points can immensely aid in forecasting trajectories. However, both of these works only consider positional information as their goal states. In our work, we propose and show the effectiveness of considering \textit{both} long-term positional goals as well as short-term intended actions.

\subsection{Intention Datasets}
To better understand agent intent in traffic scenes, a few works have proposed datasets that contain intention labels to study underlying intent in addition to the traditional trajectory prediction task. 
The JAAD~\cite{rasouli2019pie}, PIE~\cite{rasouli2017they} and STIP~\cite{liu2020spatiotemporal} datasets are recent datasets designed to study pedestrian intent. The JAAD dataset focuses on traffic scene analysis and behavior understanding of pedestrian at intersection crossing scenarios. The PIE dataset expands on JAAD further and contains more annotations for both intention estimation and trajectory prediction. PIE \cite{rasouli2019pie} only predicts intent at the current timestep and focuses on shorter horizon predictions (1.5 seconds). The STIP dataset solves the limitation of only being able to do single-shot intention prediction, as it contains frame-wise intention labels for up to 3 seconds. However, this dataset only contains "crossing/not crossing" labels for pedestrians and does not focus on trajectory prediction. All these datasets only consider intentions of pedestrians at intersections which may not capture the intents of all agents in a highly complex traffic environment with both vehicles and pedestrians.

IntentNet \cite{casas2018intentnet} does consider intents for vehicle trajectory prediction; however, they do not consider frame-wise intentions. Furthermore, the dataset and labels are not publicly available. TITAN~\cite{malla2020titan} is another driving action dataset collected from egocentric view. Although it can be potentially used for intention prediction of traffic agents, it only contains ego-view tracklets and lacks environmental and LiDAR information that can be crucial to find agents' intent. Both these works also only focus on short term predictions (less than 3 seconds). Compared with general driving dataset (such as Waymo \cite{Sun_2020_CVPR}, Nuscenes\cite{Caesar_2020_CVPR}, and INTERACTION\cite{interactiondataset}), LOKI extends the standard bounding box, track id, etc. to richer intention, contextual and environmental labels.

To the best of our knowledge, currently no publicly available dataset contains detailed, frame-wise annotations to allow for heterogeneous multi-agent trajectory forecasting \textit{and} intention prediction in joint camera and lidar space. Our dataset contains very diverse traffic scenarios through long data collection periods in different locations, weather conditions, roads and lighting. Table~\ref{tab:datasets_comparison_1} shows the details of our LOKI dataset in comparison to other recently available intention datasets (PIE, JAAD, STIP).

\noindent
\section{LOKI Dataset}
Exploring predictions in a large traffic environment is a complex problem because the future behavior of each traffic participant is not only indicated by the past behavior, but also highly impacted by the future goals and intentions. With a lack of comprehensive benchmarks for this purpose, we introduce a large scale dataset that is designed for the task of joint intention and trajectory prediction. Our dataset is collected from central Tokyo, Japan using an instrumented vehicle that is equipped with a camera \if(SEKONIX SF332X-10X)\fi, LiDAR \if(Velodyne VLP-32C)\fi, GPS and vehicle CAN BUS. The recordings are suburban and urban driving scenarios that contain diverse actions and interactions of heterogeneous agents, captured from different times of the day.

From our recordings, we extracted 644 scenarios with average 12.6 seconds length. The synced LiDAR data and RGB image were down sampled to 5HZ for annotation. The total number of agents is over 28K including 8 classes (\textit{i.e.}, Pedestrian, Car, Bus, Truck, Van, Motorcyclist, Bicyclist, Other) of traffic agents, which results in 21.6 average agents in a scene. We annotated all these agents' bounding boxes (total 886K) in the RGB image (2D) as well as LiDAR point cloud (3D) by linking with a same track-ID. The comparison with existing benchmarks is shown in Table~\ref{tab:datasets_comparison_1}. The LOKI dataset is annotated with unique attributes that can influence agents' intent such as interaction related labels, environmental constraints and contextual information.

\begin{table}[t!]
\begin{center}
   \resizebox{0.48\textwidth}{!}
   { \begin{tabular}{l||c|c|c|c}
         &PIE~\cite{rasouli2019pie}&JAAD~\cite{rasouli2017they}&STIP~\cite{liu2020spatiotemporal}&LOKI (ours) \\\hline
         $\#$ of scenarios&-&346&556&\textbf{644}\\
         $\#$ of agents&1.8K&2.8K&3.3k&\textbf{28K}\\
         $\#$ of labeled agents&1.8K&0.6K&3.3&\textbf{28K}\\
         $\#$ of classes&1&1&1&\textbf{8}\\
         $\#$ of bboxes&740K&391K&350k&\textbf{886K}\\
         $\#$ of agent types&1 (Ped)&1 (Ped)& 1 (Ped)&\textbf{8 classes}\\
         Avg. agent per frame&2.5&5.2&3.2&\textbf{21.6}\\
         Annotation freq.&-&-&2 FPS&\textbf{5 FPS}\\
         Frame-wise labels&no&\cmark&\cmark&\cmark\\
         RGB Images &\cmark&\cmark&\cmark&\cmark\\
         LiDAR Point cloud  &no&no&no&\cmark\\
         2D Bounding box&\cmark&\cmark&\cmark&\cmark\\
         3D Bounding box&no&no&no&\cmark\\
         Lane Info&no&no&no&\cmark\\
         Pedestrian attributes&no&\cmark&no&\cmark\\
         \hline

    \end{tabular}}
    \caption{ Comparison of LOKI dataset with PIE~\cite{rasouli2019pie}, JAAD~\cite{rasouli2017they} and STIP~\cite{liu2020spatiotemporal}.}
     \vspace{-0.6cm}
    \label{tab:datasets_comparison_1}
\end{center}
\end{table}

\begin{figure}[t!]
\begin{center}

    \includegraphics[width=0.48\textwidth]{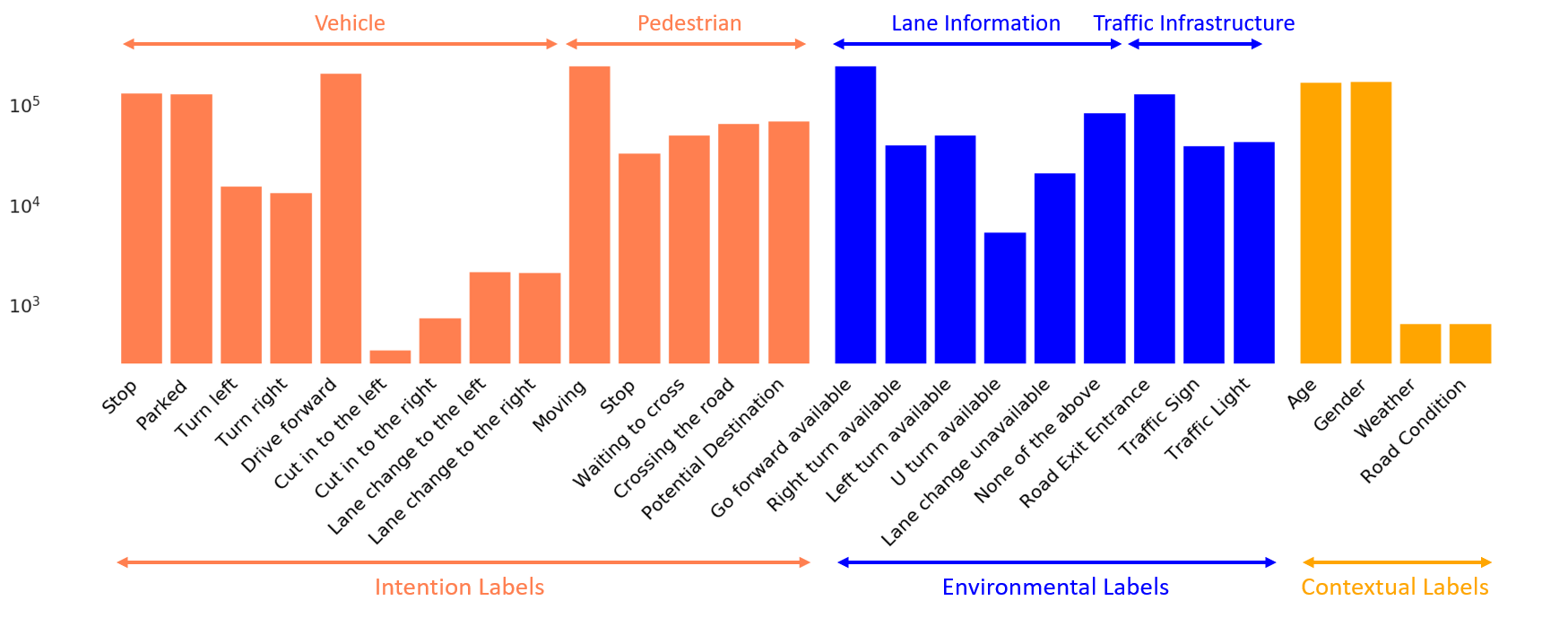}
    \caption{Distribution of labels sorted according to the different types of intention among the different classes}
     \vspace{-0.6cm}
    \label{fig:flow_chart}
\end{center}
\end{figure}

\subsection {Dataset Annotation}
Considering that LiDAR point clouds better capture positional relations among agents than RGB images, we annotate 3D bounding box of agents with their orientation, potential destination of pedestrians, road entrance / exit, and agents' intention as well as action labels in this space. In contrast, in the RGB image space we leverage its contextual clarity to annotate environmental labels such as lane information (what actions can be made from this lane), lane number for vehicles (relative position with respect to the autonomous agent), the gender and age for pedestrian, the state of traffic light, and the type of traffic sign. Note that we also annotate 2D bounding box, potential destination and road entrance / exit information in the RGB space to inspire the potential research in the egocentric view. By using the consistent tracking ID between the same agent in the 3D LiDAR space and 2D image space, our labels can be shared across different spaces.

\begin{figure}[t!]
\begin{center}

    \includegraphics[width=0.45\textwidth]{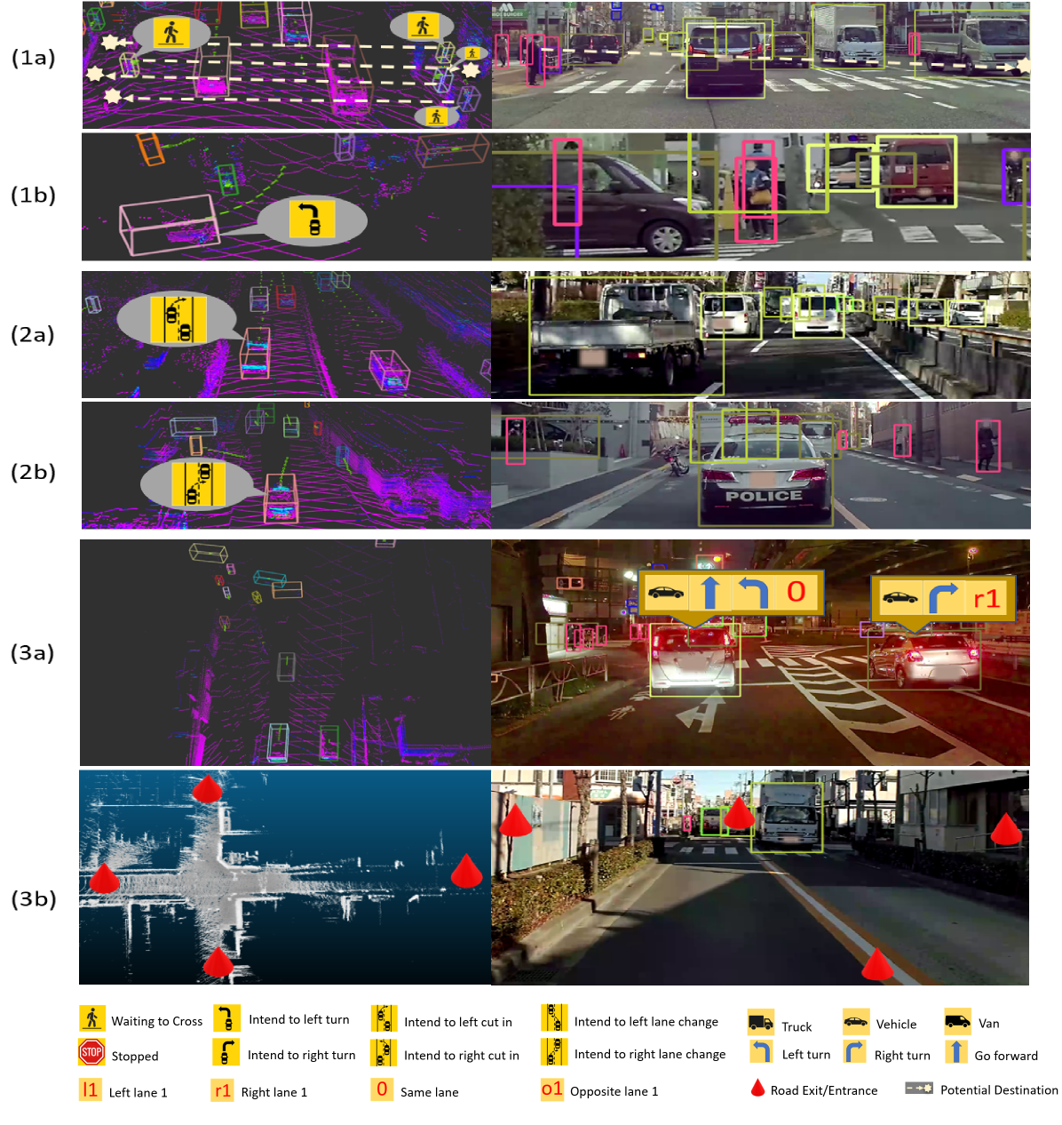}
    \caption{Visualization of three types of labels: (1a-1b) Intention labels for pedestrian; (2a-2b) Intention labels for vehicle; and (3a-3b) Environmental labels. The left part of each image is from laser scan and the right part is from camera. In (1a), the current status of pedestrian is "Waiting to cross", and the potential destination shows the intention of pedestrian. In (3a), the blue arrow indicates the possible action of the current lane where the vehicle is on, and the red words present the lane position related to the ego-vehicle.}
     \vspace{-0.6cm}
    \label{fig:flow_chart_3types}
\end{center}
\vspace{-0.3cm}
\end{figure}

To dig into more complex prediction researches, our dataset provides denser agents per frame and more meticulous intention attributes compared to other datasets.

We have three types of labels in the LOKI dataset: Intention labels, Environmental labels and Contextual labels to explore how these can affect the future behavior of agents (details and visuals are in Figure \ref{fig:flow_chart} and Figure \ref{fig:flow_chart_3types}).

\noindent
\textbf{Intention labels} Intentions are defined to be "how" an actor decides to reach a goal via a series of actions \cite{rasouli2021pedestrian}. At each frame, we annotated the current actions of the traffic participants and then used future actions to generate our intention labels. For example, if the current action of vehicle is “Moving” and the future action in 1 second is "Stopped", the vehicle's current intention is to stop. Various intention horizons can be explored; we use $0.8s$, as we explore how short-term intent can help guide trajectory prediction.

\noindent
\textbf{Environmental labels} The environment of driving scene can heavily impact the intention of agent especially for the driving area users, so we include the environmental information such as "Road Exit and Entrance" positions, "Traffic light", "Traffic Sign", "Lane Information" in the LOKI dataset. Those labels determined by the structure of the road and the traffic rules that can be applied to any agent in the scene. The lane information includes the allowed actions of the current lane where the vehicle is on and the relative position between other vehicle and ego-vehicles.

\noindent
\textbf{Contextual labels} There are some other factors may also affect the future behavior of agent. We define the "Weather", "Road condition", "Gender", "Age" as external contextual labels. These factors are the characters of the agent or environment which can cause the different intentions even under similar environment condition.

\section{Proposed Method}

\subsection{Problem Formulation}
In this work, we tackle the problem of multi-agent trajectory forecasting while concurrently predicting agent intentions. The type of intentions vary between agent classes: vehicles and pedestrians. We formulate the problem as follows. Suppose in a given scene, $\mathcal{S}$, we have $N$ agents, $A_{1:N}$. Given the past $t_{obs}=3s$ of trajectory history in BEV coordinates, the problem requires forecasting the future $t_{pred}=5s$ coordinates of the agent in top-down image space. Since our dataset allows for frame-wise intention predictions depending on agent type (pedestrians vs. vehicles), we define another task to predict discrete intentions for each agent at each timestep, in addition to the traditional trajectory prediction problem.

\subsection{Model Design}

\subsubsection{Long-term Goal Proposal Network}
Intuitively, agents have a predetermined, long-term goal that they want to reach. Many recent goal-directed works have focused on modeling this through estimating final "endpoint" or "goal state" distributions as done in \cite{mangalam2020not, mangalam2020goals, zhao2020tnt, deo2020trajectory, choi2019drogon}. Inspired by agents' rational decision-making process and the success of prior works, we design a goal network similar to the method proposed in \cite{mangalam2020not}. For each agent, $A_k$, we use a Conditional Variational Autoencoder (CVAE) to estimate the final long term goal $G_{k}$ that is simply the estimated position in BEV $u_{k_f}=(x_{k_f}, y_{k_f})$ where $f$ indicates the final frame. The inputs into the CVAE are the encodings from the Observation RNN Encoder. The goal network only consider agents' own  history, as agents have a predetermined long term goal irrespective of other agents.

\begin{figure}[t!]
\begin{center}
    \includegraphics[width=0.48\textwidth]{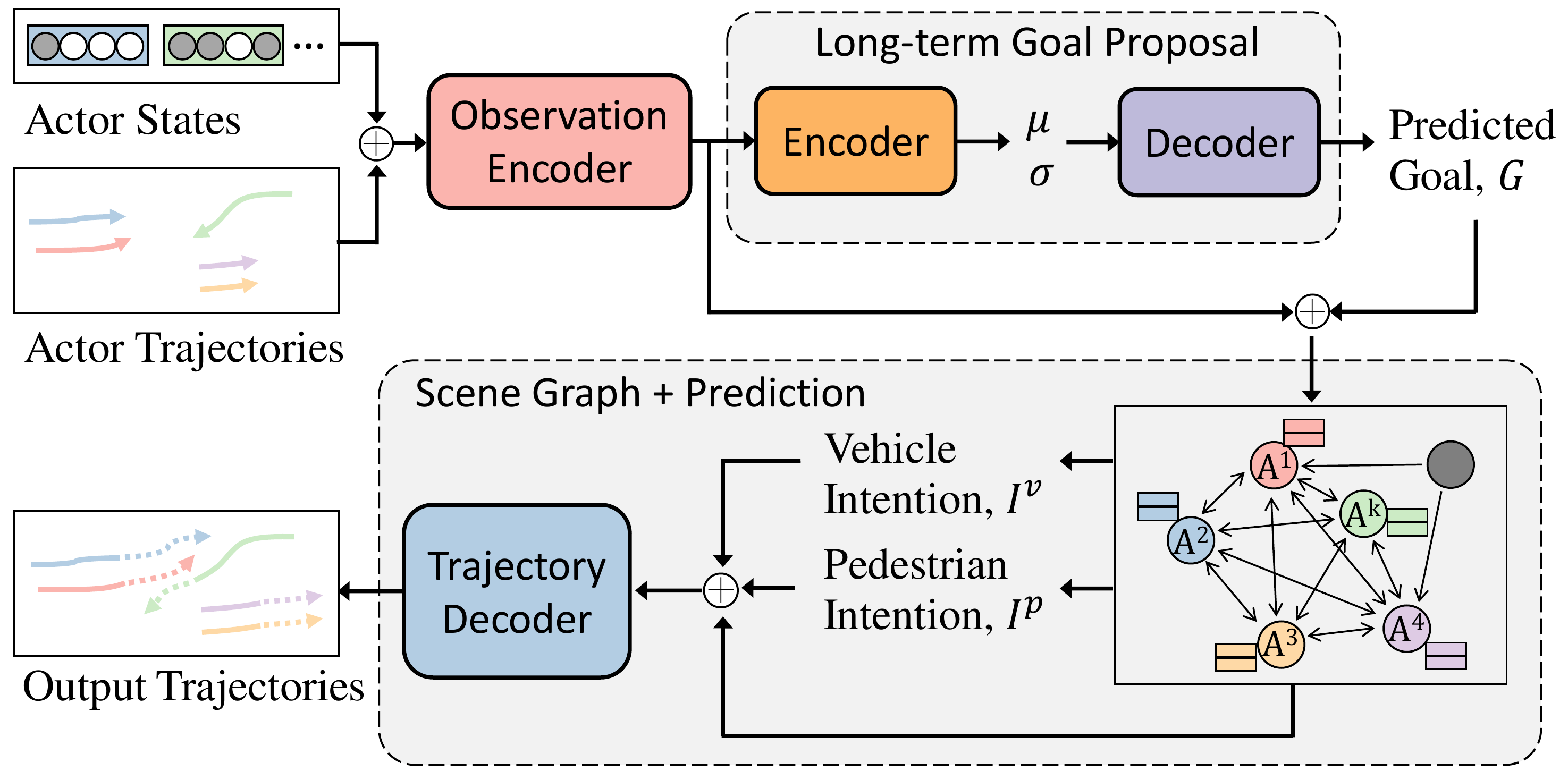}
    \caption{Our model first encodes past observation history of each agent to propose a long-term goal distribution over potential final destinations for each agent independently. A goal, $G$ is then sampled and passed into the Joint Interaction and Prediction module. A  scene graph is constructed to allow agents to share trajectory information, intentions, and long-term goals. Black nodes denote road entrance/exit information which provides agents with map topology information. At each timesteps, current scene information is propagated through the graph. We then predict an intent (the action will the agent take in the near future) for each agent. Finally, the trajectory decoder is conditioned on predicted intentions, goals, past motion, and scene before forecasting the next position. This process is recurrently repeated for the horizon length. Note that $\bigoplus$ denotes concatenation.
}
     \vspace{-0.6cm}
    \label{fig:model}
\end{center}
\end{figure}

\subsubsection{Scene Graph + Trajectory Decoder}

Our main insight and promising directions from our proposed dataset comes from agents' short-term intentions. As described earlier, we have different intentions for pedestrians and vehicles. Without loss of generality, we will refer to agents, $A$, and intentions, $I$, without specifying the type of agent. We believe agents' have intermediate stochastic intents that can change depending on agent behavior, agent-agent interaction, or environmental factors. To account for this, we construct a traffic scene graph $\mathcal{G}$ to account for social and environmental factors that may affect intent and trajectory prediction.

More concretely, suppose we have a scene graph $\mathcal{G}=(\mathcal{V},\mathcal{E})$ where vertices, $\mathcal{V}$, denote agents and road entrances/exits and edges, $\mathcal{E}$, capture agent-agent and agent-map influence. In a given scene, for neighboring agents $v_i$ and $v_j$, there is a directional edge $e_{ij}$ if agent $i$ affects agents $j$ (within a certain distance threshold). Static road entrance/exit nodes can affect agents but without incoming edges. We connect a directional edge $e_{ij}$ if road entrance/exit node $i$ is within a certain distance from agent $j$. 

We then predict agents' future locations via a daisy chained process described as follows. At each frame, $m$, our model first shares information between agents via the attention mechanism used in \cite{transconv}:
\begin{align*}
    x_i^{t+1} = \gamma(x_i^t) + \sum_{x_j \in \mathcal{N}(x_i)} \alpha_{ij} * \phi(x_j^t, e_{ij}),
\end{align*}
where $x_i^{t+1}$ represents the updated node features following attention-based feature aggregation with all of its neighbors $x_j \in \mathcal{N}(x_i)$. We use agents' velocities and relative positions as edge features. These features are encoded by a 2-layer MLP prior to message passing at each timestep. We use the scaled dot-product attention \cite{transconv} formulation:
\begin{equation*}
a_{ij}={{softmax}}(\frac{\psi(x_i)^T\xi(x_j, e_{ij})}{\sqrt{d}})
\end{equation*}
Here, $a_{ij}$ represents the attention coefficient between two nodes $i$ and $j$ and $d$ represents the degree of the node. We use a single-layer  for $\phi$, $\gamma$, $\psi$, and $\xi$. 

After message passing which allows agents to share their past trajectory, goal, and intention information along with road information through the road entrance/exit nodes, our model then predicts agent intent, which we define to be the agent's future action $m+q$ frames ahead. In our experiments, we set $q=4$, thus predicting short-term intent $0.8s$ in the future. We then condition trajectory prediction for frame $m+1$ based on agent intent at frame $m$. This process of information sharing and intention conditioning is recurrently repeated for the next $f-ob$ timesteps where $f$ denotes the last prediction frame number and $ob$ denotes the last observation frame. Formally, at each frame, $m$, we first estimate the probability distribution over a discrete set of intentions (different set of intentions for pedestrian vs. vehicle) for an agent, $A_i$:
\begin{align*}
P(I_{i_m}|I_{i_{ob:{m-1}}}, U_{i_{0:{m-1}}}, G_{i}, a_{i_{0:{ob}}}, \cup_{A_j \in \mathcal{N}(A_i)} I_{j_{ob:{m-1}}}, \\
U_{j_{0:{m-1}}}, G_{j}, a_{j_{0:{ob}}}, R_{ee})
\end{align*}
where $I$ refers to intention, $U$ is position, $G$ is long-term positional goal, $a$ is action, and $R_{ee}$ refers to road entrances/exit labels. The intention networks are two-layer MLPs which predicts intention using each actor's updated hidden states from the most recent message passing. Following this, we then predict the next position of each agent, $U$, conditioned as follows:
\begin{align*}
P(U_{i_{m+1}}|I_{i_{o:{m}}}, U_{i_{0:{m}}}, G_{i}, a_{i_{0:{ob}}}, \cup_{A_j \in \mathcal{N}(A_i)} I_{j_{o:{m}}}, \\ 
U_{j_{0:{m}}}, G_{j}, a_{j_{0:{ob}}}, R_{ee})
\end{align*}
The trajectory decoder module consists of a GRU that updates each actor's current hidden state followed by a 2-layer MLP used to predict positions at each step. The overview of our model is illustrated in Figure \ref{fig:model}. Specific model architecture details will be provided in the supplementary material.

\subsubsection{Loss Functions}
Our goal proposal network (GPN) follows the approach introduced in \cite{mangalam2020not} and is trained via the following loss function:
\begin{equation*}
\mathcal{L}_{GPN} = \alpha_1
D_{KL}(\mathcal{N}({\mu}, {\sigma}) \Vert \mathcal{N}(0, {I})) + \alpha_2 \Vert\hat{G} - {G}\Vert^2_2
\end{equation*}

Here $\alpha_1$ and $\alpha_2$ are tunable parameters to weight the KL Divergence loss and goal reconstruction loss for training the CVAE. We observed that training via conditioning with ground-truth goal positions helps with model convergence because the intentions are dependent on the long-term goal.

Our decoder module which is responsible for both intention and trajectory prediction is composed of separate loss terms for each. Our intention loss is defined as follows:
\begin{equation*}
\mathcal{L}_{int} = -\sum_{j=t_{ob}+1}^{t_f}\sum_{i=0}^{n} w_{ij}*y_{ij}*log(\hat{y_{ij}})
\end{equation*}

Due to heavy class imbalance, we not only augment rare trajectories such as lane changes and turning but also weight the cross entropy loss by $w_i$, which is the inverse frequency of the class.

Since we predict offsets in position (velocity) rather than position directly for better model convergence, our loss is on the predicted velocity $V$ for all timesteps:
\begin{equation*}
\mathcal{L}_{traj} = ||V-\hat{V}||_2
\end{equation*}

We train our network end-to-end by weighting each of the loss terms:

\begin{equation*}
\mathcal{L}_{Final} = \lambda_1 \mathcal{L}_{GPN} + \lambda_2\mathcal{L}_{int} + \lambda_3\mathcal{L}_{traj}
\end{equation*}

\subsubsection{Evaluation Metrics}
For trajectory prediction evaluation, we use the standard Average Displacement Error (ADE) and Final Displacement Error (FDE) metrics:

\begin{multicols}{2}
  \begin{equation*}
    \textit{ADE} = \frac{\sum_{j=t_{ob} + 1}^{t_f}\Vert \hat{{u}}_{j} - {u}_{j} \Vert_2} {(t_{f}-t_{ob})}
  \end{equation*}\break
  \begin{equation*}
    \textit{FDE} = \Vert \hat{{u}}_{t_f} - {u}_{t_f} \Vert_2 
  \end{equation*}
\end{multicols}

where $\hat{{u}}$ and ${u}$ are the estimated and ground truth positions respectively. Furthermore, we use the $minADE$-$N$ and $minFDE$-$N$ error metric introduced in \cite{gupta2018social} for multimodal evaluation. The metric is simply the minimum ADE and FDE out of $N$ future trajectories predicted at test-time.

For intention prediction, we evaluate frame-wise classification accuracy of intents and visualize the confusion matrix to analyze classification performance.

\section{Experiments}

\begin{table*}[t]
\centering
\vspace{-0.1cm}
\resizebox{0.75\textwidth}{!}{
\begin{tabular}{|c|c|c|c|c|c|c|c|c|c|c|c|c|}
\hline
                       & \multicolumn{2}{c|}{S-STGCNN}    & \multicolumn{2}{c|}{EvolveGraph} & \multicolumn{2}{c|}{PECNet}     & \multicolumn{2}{c|}{Ours}       & \multicolumn{2}{c|}{Ours + IC}           & \multicolumn{2}{c|}{Ours + IC + SG}      \\ \hline
\multicolumn{1}{|l|}{} & \multicolumn{1}{l|}{ADE} & FDE   & \multicolumn{1}{l|}{ADE} & FDE  & \multicolumn{1}{l|}{ADE} & FDE  & \multicolumn{1}{l|}{ADE} & FDE  & \multicolumn{1}{l|}{ADE} & FDE           & \multicolumn{1}{l|}{ADE} & FDE           \\ \hline
Pedestrians            & 0.96 & 1.98  & 0.83 & 1.49 & 0.79 & 1.31 & 0.61 & 1.38 & 0.56 & 1.24 & \textbf{0.55} & \textbf{1.21} \\ \hline
Vehicles               & 3.03 & 7.01  & 2.58 & 6.63 & 2.52 & 6.34 & 2.37 & 6.20 & \textbf{2.23} & \textbf{5.80} & 2.24  & 5.82 \\ \hline
Lane Change            & 4.41 & 10.17 & 2.96 & 7.92 & 2.78 & 7.60 & 2.93 & 7.88 & \textbf{2.47}  & 6.78 & 2.52 & \textbf{6.71} \\ \hline
Turn                   & 3.48 & 8.15  & 3.13 & 7.85 & 2.97 & 7.44 & 2.76 & 7.26 & 2.69 & 7.03 & \textbf{2.69} & \textbf{7.02} \\ \hline
\end{tabular}}
\vspace{-0.2cm}
\caption{\small{\textbf{Trajectory error metrics for N=1 samples}: ADE and FDE of various state-of-the-art baselines and our method using unimodal (single-shot) evaluation. Reported errors are in meters. Lower is better. We show results evaluated on separate classes to gain more insight on prediction performance. We report errors on  1) pedestrians, 2) vehicles (non-static), 3) agents that change lanes, and 4) agents that turn}.}
\label{tab:unimodal}
\end{table*}

\begin{table*}[t]
\centering
\resizebox{0.85\textwidth}{!}{
\begin{tabular}{|c|c|c|c|c|c|c|c|c|c|c|c|c|c|c|}
\hline
            & \multicolumn{2}{c|}{S-GAN} & \multicolumn{2}{c|}{S-STGCNN} & \multicolumn{2}{c|}{EvolveGraph} & \multicolumn{2}{c|}{PECNet} & \multicolumn{2}{c|}{Ours} & \multicolumn{2}{c|}{Ours + IC} & \multicolumn{2}{c|}{Ours + IC + SG} \\ \hline
            & ADE          & FDE         & ADE           & FDE           & ADE          & FDE           & ADE          & FDE          & ADE         & FDE         & ADE            & FDE           & ADE              & FDE              \\ \hline
Pedestrians & 1.04         & 2.18        & 0.63          & 1.01          & 0.55         & 0.79         & 0.51         & 0.70         & 0.36        & 0.70        & 0.37           & 0.71          & \textbf{0.34}    & \textbf{0.64}    \\ \hline
Vehicles    & 3.57         & 8.05        & 2.28          & 4.46          & 1.72         & 3.41         & 1.59         & 3.05         & 1.33        & 3.09        & 1.20           & 2.63          & \textbf{1.18}    & \textbf{2.64}    \\ \hline
Lane Change & 3.50         & 8.41        & 3.00          & 6.09          & 1.86         & 3.39         & 1.62         & 2.85         & 1.42        & 3.30        & 1.26           & 2.70          & \textbf{1.22}    & \textbf{2.71}    \\ \hline
Turn        & 3.75         & 9.01        & 2.68          & 5.71          & 2.25         & 4.32         & 1.96         & 4.07         & 1.54        & 3.59        & 1.45           & 3.24          & \textbf{1.40}    & \textbf{3.13}    \\ \hline
\end{tabular}}
\vspace{-0.2cm}
\caption{ \small{\textbf{Trajectory error metrics for N=20 samples}: ADE and FDE of various state-of-the-art baselines and our method using multimodal evaluation. Reported errors are in meters. Lower is better. We report errors on the same classes described in Table \ref{tab:unimodal}.}
}
\label{tab:multimodaltable}
\end{table*}

In this section, we present results of our model on trajectory \& intent prediction tasks and demonstrate a superior performance against prior state-of-the-art baselines (with publicly available code) across a variety of settings. We benchmark against PECNet \cite{mangalam2020goals}, a strong scene agnostic trajectory prediction method with state-of-the-art performance on standard intention agnostic prediction datasets. S-STGCNN \cite{mohamed2020social} and S-GAN \cite{gupta2018social} are strong socially-aware models that achieved prior state-of-the-art on various benchmarks. We also report an interesting ablation on the effect of annotation frequency on the final performance, which confirms our hypothesis for the effectiveness of detailed intent annotations in trajectory prediction.

\paragraph{Trajectory Prediction Performance.}
We report our model's performance and benchmark it against prior state-of-the-art models for unimodal (single shot, N = 1) prediction in Table \ref{tab:unimodal} and for multimodal predictions (N = 20 shots) in Table \ref{tab:multimodaltable}.
Our ablations are with Ours (without action/intention labels), IC (with action/intention labels for intention conditioning), SG (with scene graph for social reasoning and environmental cues).

Several interesting trends emerge. First, we observe that in the single shot setting, our intention conditioned model outperforms prior state-of-the-art method by a significant margin of 12\% in ADE, 9\% in FDE. Second, we see a similar trend in multi-shot prediction setting as well with our model outperforming PECNet by 33\% in ADE and 9\% in FDE for pedestrians and a delta of 26\% in ADE and 13\% in FDE for moving vehicles. Third, notice that the performance gap is significant in hard non-linear cases such as lane changes and turns, where our model achieves 30\% and 16\% better performance in ADE and FDE respectively.

Also noteworthy is the crucial effect of  conditioning predictions on intentions and incorporating social and environmental cues through the scene graph, which is also shown in Table \ref{tab:unimodal} and Table \ref{tab:multimodaltable}. We note that both intention cues and scene graph information are critical to overall performance, with intention improving ADE performance by up to 7\% and 8\% across all agent types (especially nonlinear trajectories such as lane changes and turns) for the unimodal and multimodal settings. We notice that the scene graph boosts performance by  3\% in ADE for the multimodal setting across all agent types. 
\begin{figure}[t]
\centering
{\includegraphics[width=0.42\textwidth]{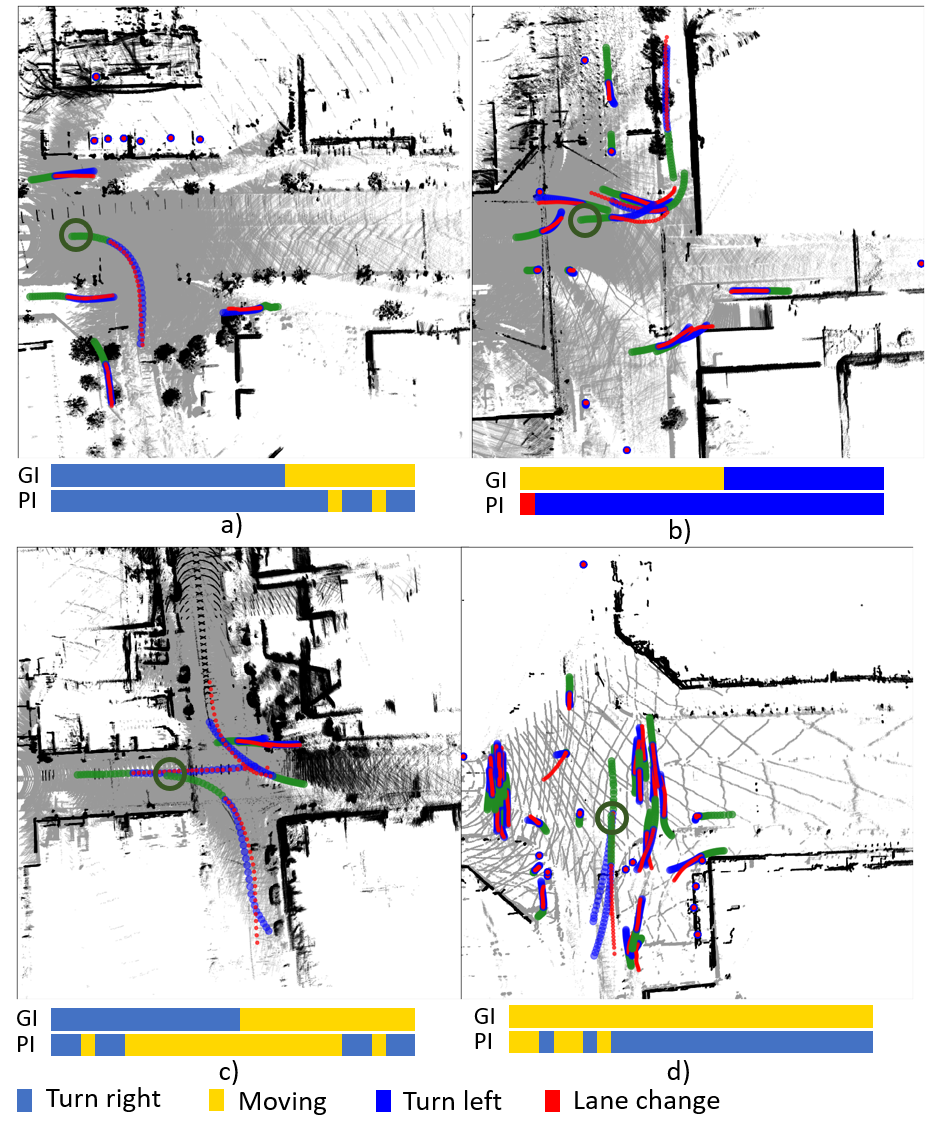}}
 \vspace{-0.2cm}
\caption{Visualization of top-1 trajectory prediction result (green: past observation, blue: ground truth, red: prediction) and frame-wise intention of a particular agent in dark green circle at the start of the observation time step(GI: Ground truth Intention, PI: Predicted Intention) is shown at the bottom of each scenario. More detailed visualizations and comparisons are provided in supplementary material. }
 \vspace{-0.25cm}
\label{fig:qualitative}
\end{figure}

\begin{figure}[t]
     \centering
     \begin{subfigure}[b]{0.22\textwidth}
         \centering
         \includegraphics[width=\textwidth]{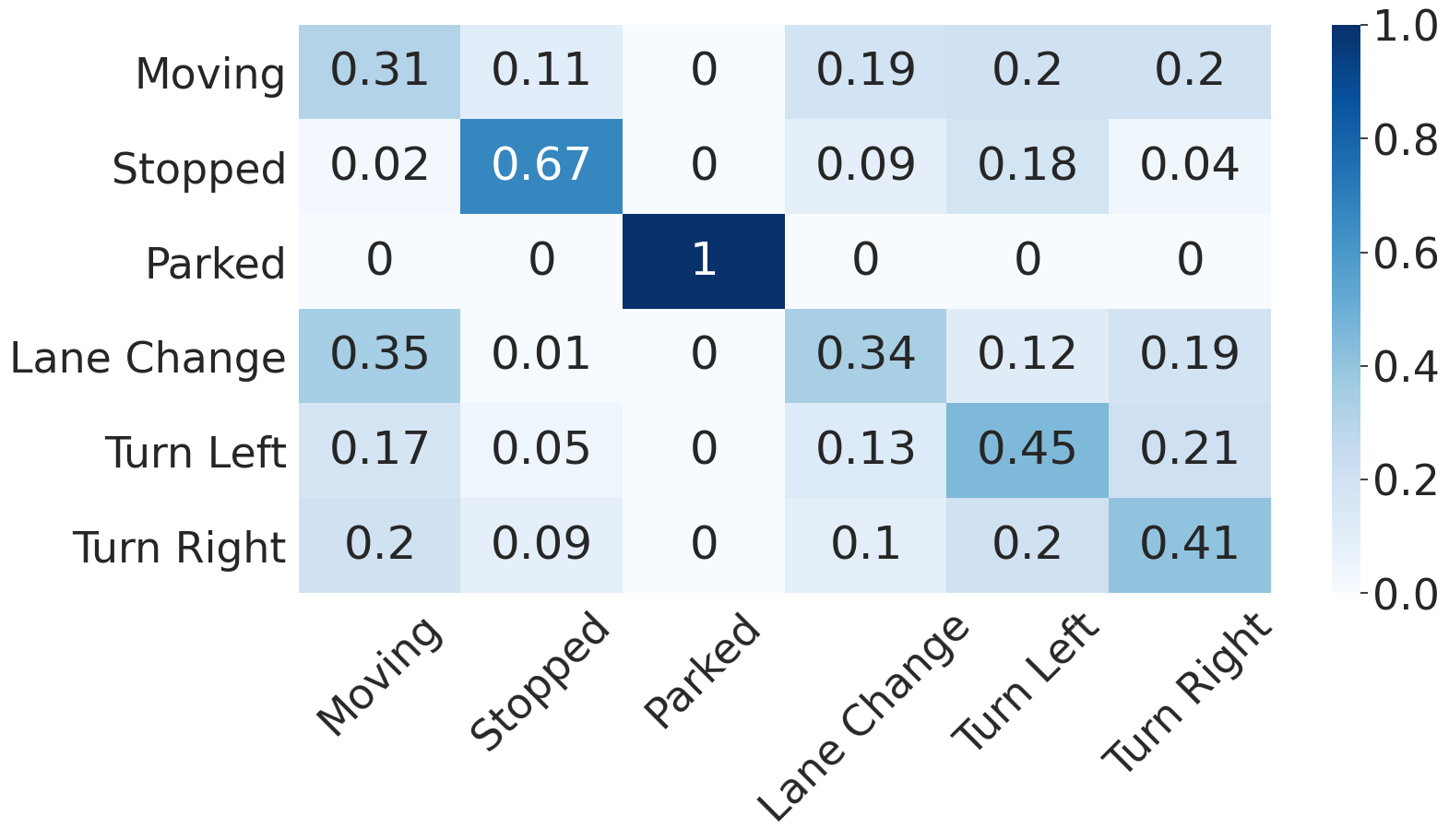}
         \caption{Vehicle (N=1)}
         \label{fig:7a}
     \end{subfigure}
     \hfill
     \begin{subfigure}[b]{0.22\textwidth}
         \centering
         \includegraphics[width=\textwidth]{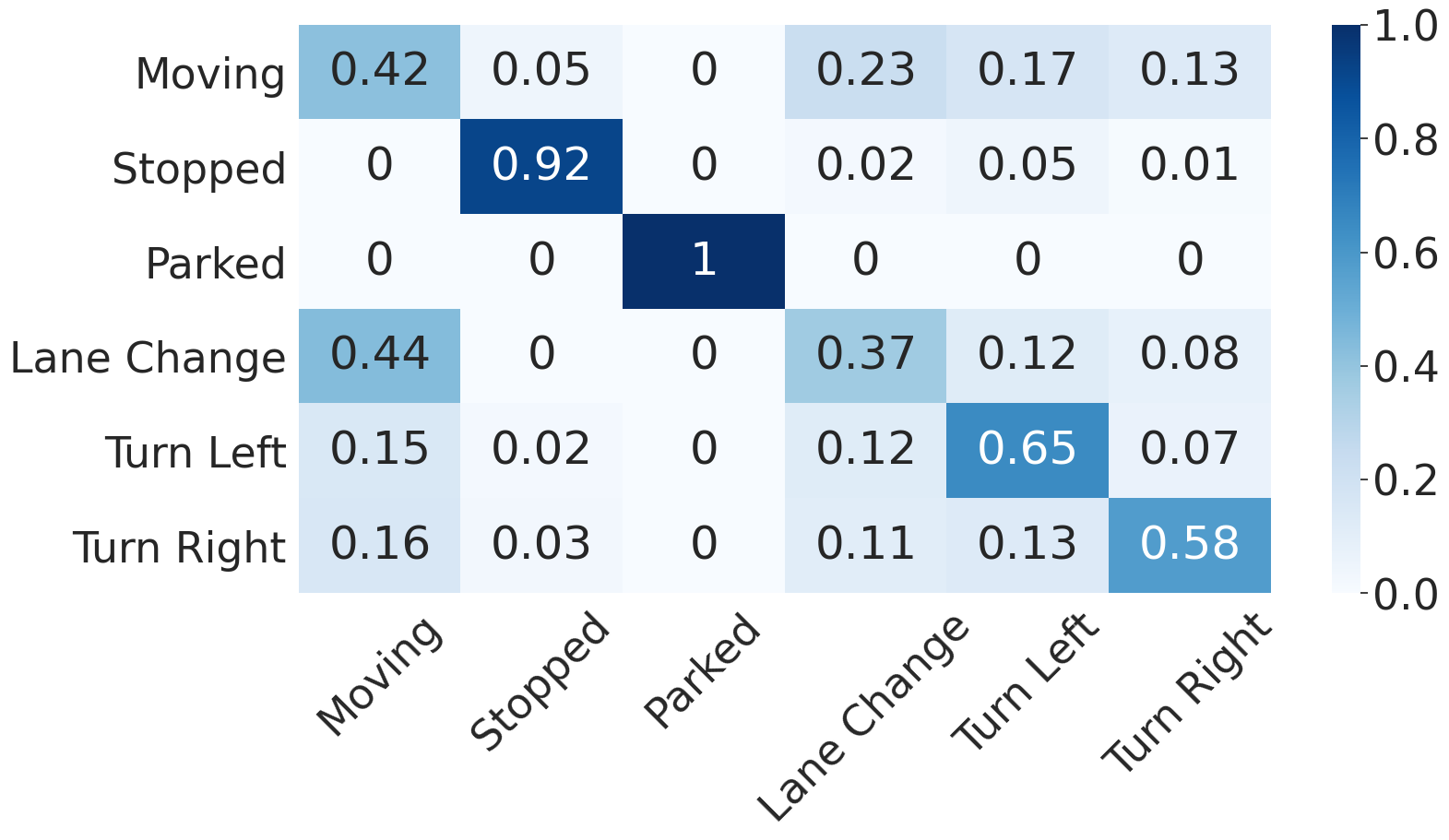}
         \caption{Vehicle (N=20)}
         \label{fig:7b}
     \end{subfigure}
    \vskip\baselineskip
     \begin{subfigure}[b]{0.22\textwidth}
         \centering
         \includegraphics[width=\textwidth]{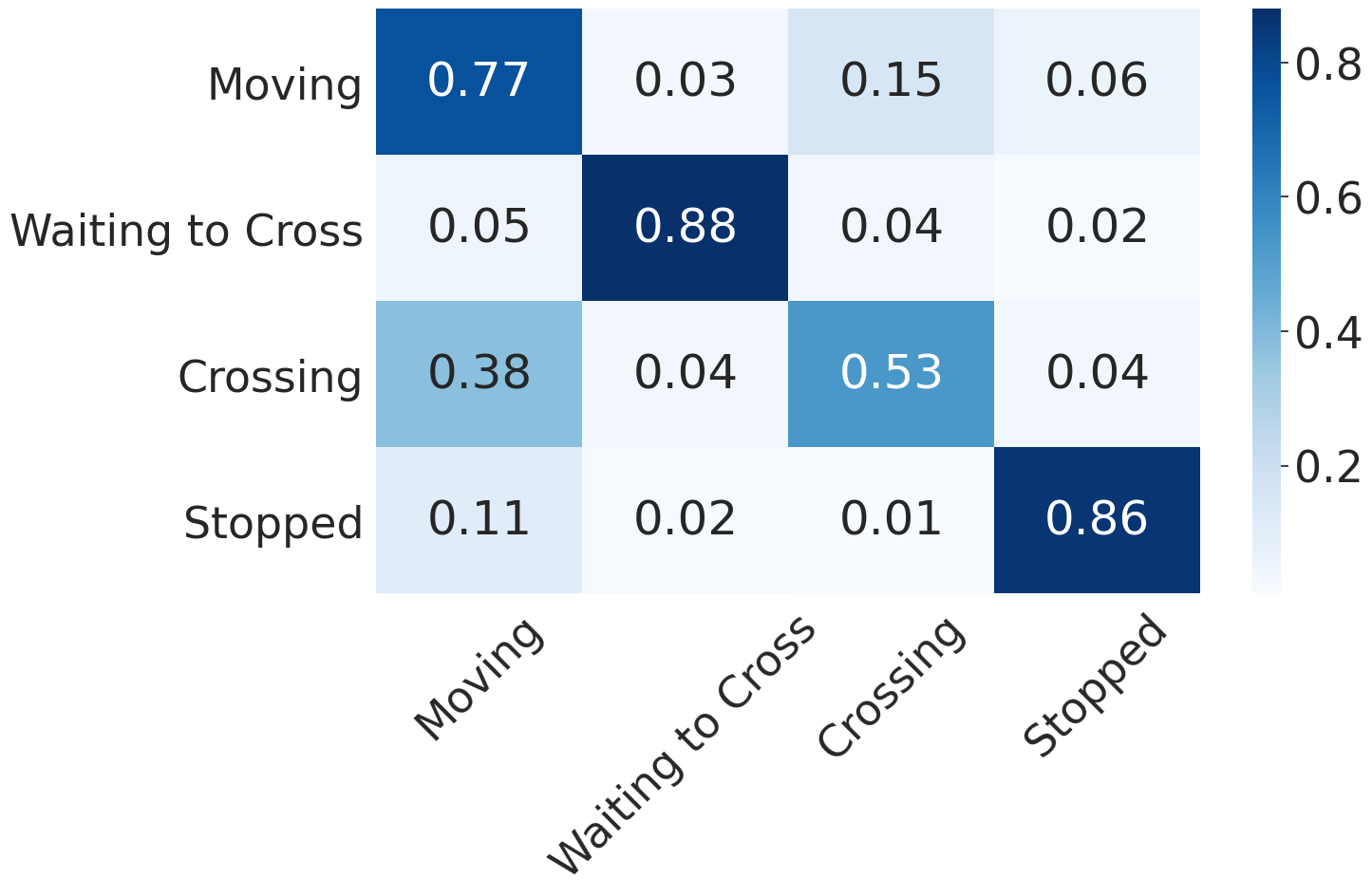}
         \caption{Pedestrian (N=1)}
         \label{fig:7c}
     \end{subfigure}
    \hfill
     \begin{subfigure}[b]{0.22\textwidth}
         \centering
         \includegraphics[width=\textwidth]{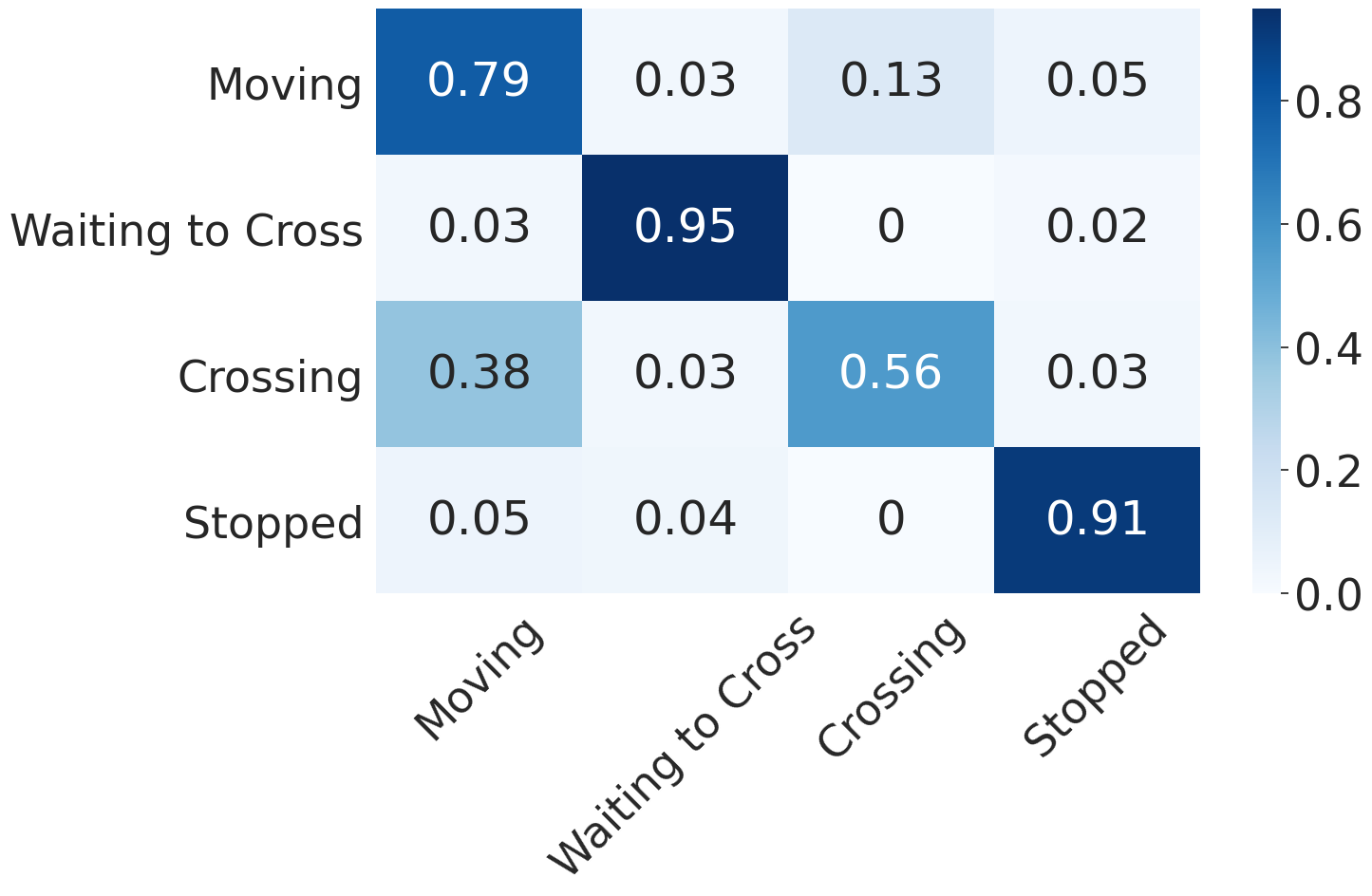}
         \caption{Pedestrian (N=20)}
         \label{fig:7d}
     \end{subfigure}

        \caption{Intention prediction confusion matrices. (a-b) results for vehicles under both unimodal and multimodal sampling, (c-d) those for pedestrians.}
         \vspace{-0.5cm}
        \label{fig:confusion_matrices}
\end{figure}

\begin{figure}[t]
\centering
{\includegraphics[width=0.9\columnwidth]{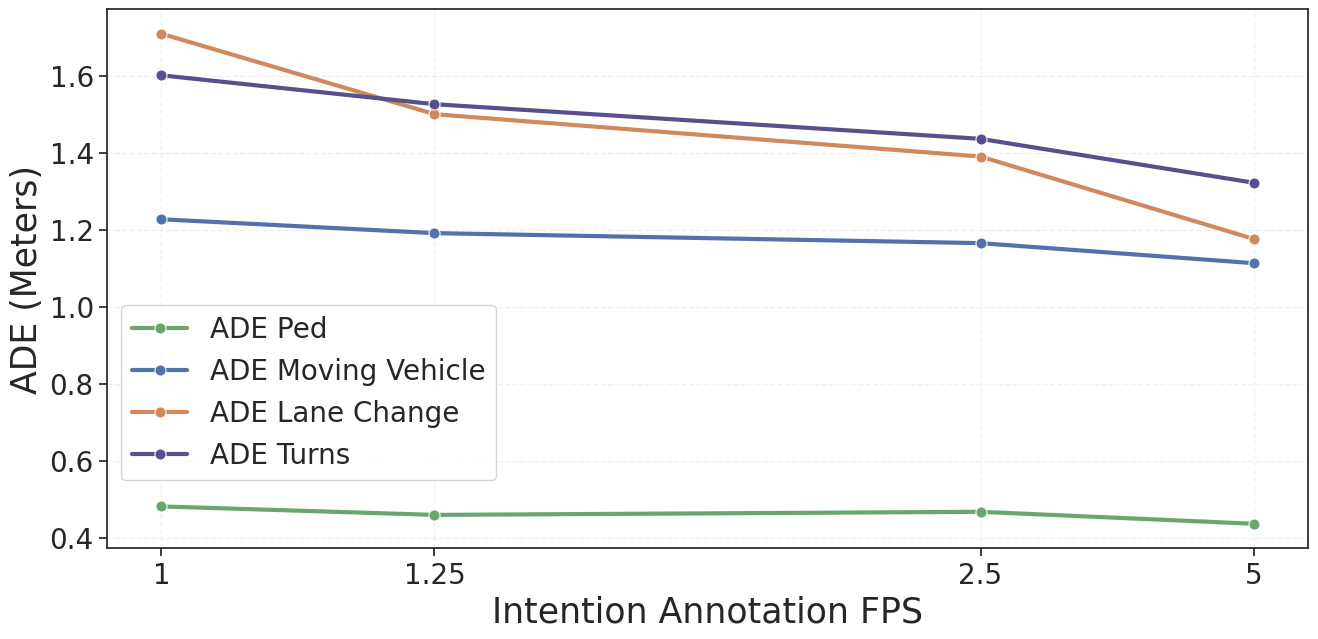}}
 \vspace{-0.2cm}
\caption{ADE Performance based on varying ground-truth intention annotation frequency.}
 \vspace{-0.51cm}
\label{fig:intfps}
\end{figure}

We notice an interesting behavior with pedestrians. Conditioning on pedestrian intent such as crossing vs. waiting to cross helps for single-shot prediction as shown in Table \ref{tab:unimodal}. However, we do not see a benefit for multimodal prediction. We hypothesize that this is because the type of intent we label for pedestrian is not as granular as for vehicles in that it does not change drastically frame-by-frame. This is validated in Figure \ref{fig:intfps} which shows experiments with downsampled intention annotations. We observe that for pedestrians, lower frequency annotations does not diminish performance as compared to vehicles due to more unconstrained behavior, we cannot have as detailed intent labels that are used for vehicles such as turn or lane change. This may explain the behavior of why intention conditioning only helps for the single-shot case for pedestrians.

In Figure \ref{fig:qualitative}, we visualize our model's best-of-20 performance. We observe that predicted trajectories are fairly accurate and with underlying turning intentions. While there are limitations in exact frame-wise intention predictions, we notice it can capture key future actions of turning and can help guide predictions.

\begin{figure}[t]
     \centering
     \begin{subfigure}[h]{0.45\columnwidth}
         \centering
         \includegraphics[width=\textwidth]{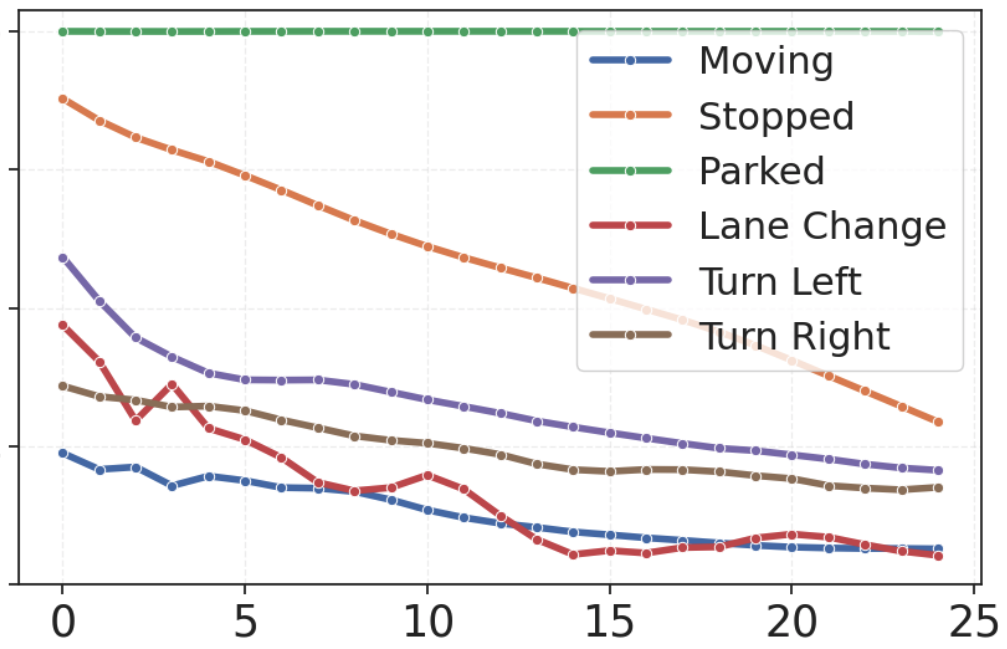}
         \caption{Vehicle (N=1)}
         \label{fig:5a}
     \end{subfigure}
     \hfill
     \begin{subfigure}[h]{0.45\columnwidth}
         \centering
         \includegraphics[width=\textwidth]{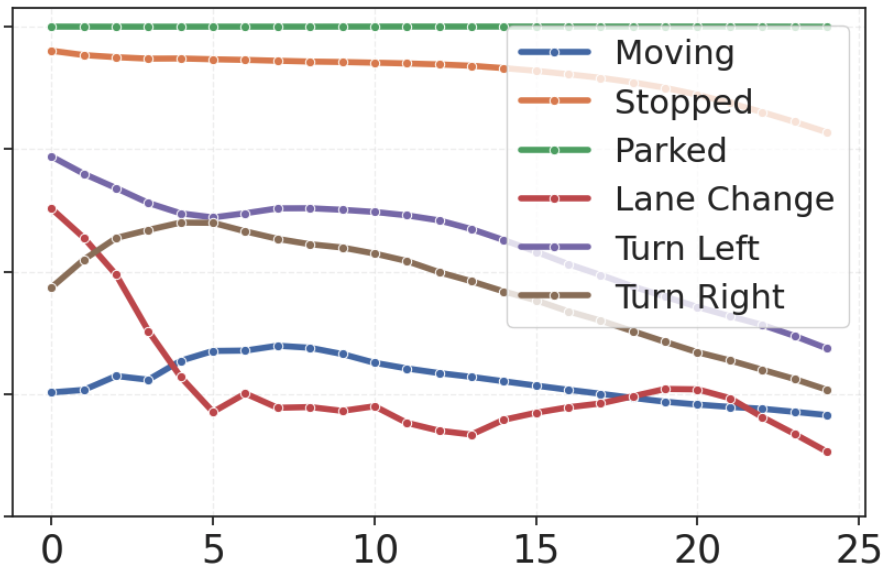}
         \caption{Vehicle (N=20)}
         \label{fig:5b}
     \end{subfigure}
    \vskip\baselineskip
     \begin{subfigure}[h]{0.45\columnwidth}
         \centering
         \includegraphics[width=\textwidth]{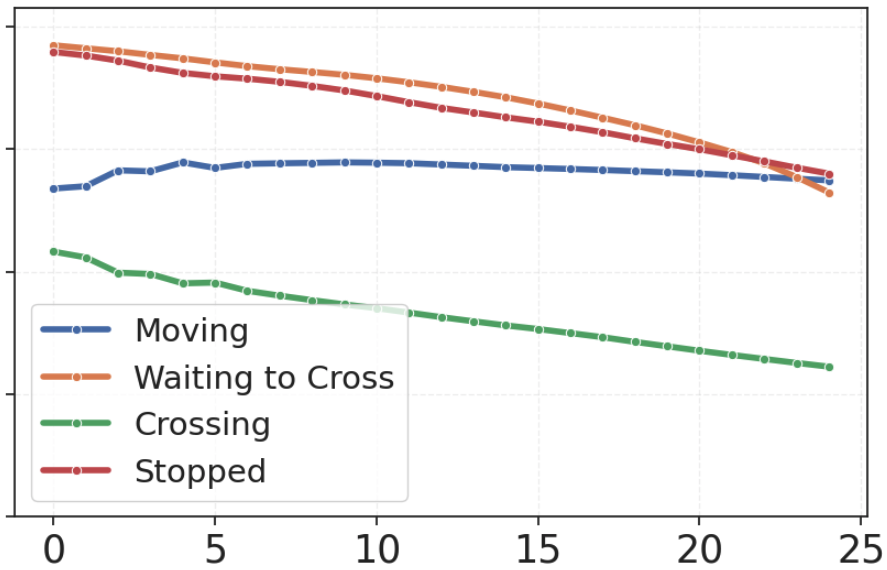}
         \caption{Pedestrian (N=1)}
         \label{fig:5c}
     \end{subfigure}
    \hfill
     \begin{subfigure}[h]{0.45\columnwidth}
         \centering
         \includegraphics[width=\textwidth]{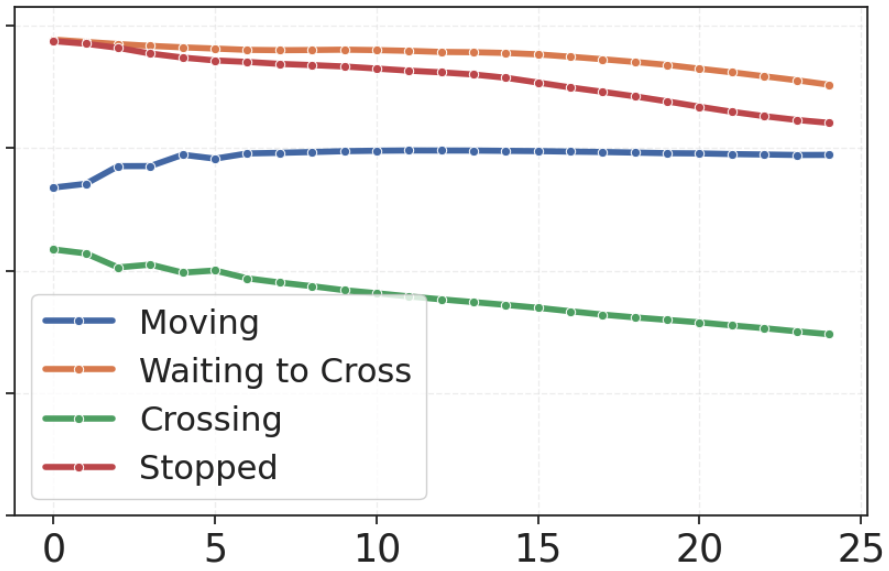}
         \caption{Pedestrian (N=20)}
         \label{fig:5d}
     \end{subfigure}
        \vspace{-0.1cm}
        \caption{\textbf{Accuracy vs. Future Horizon (in frames).} The x axis of each figure is time and the y axis of each figure is accuracy (from 0 to 1). The change of intention prediction accuracy over a time horizon for both unimodal and multimodal predictions. In (a-b) we plot intention accuracy over time for vehicles for N=1 and N=20 samples respectively. In (c-d) we plot intention accuracy over time for pedestrians with N=1 and N=20 samples.}
         \vspace{-0.6cm}
        \label{fig:accuracyhorizon}
\end{figure}

\noindent \textbf{Intention Prediction}:
In addition to trajectory prediction, our dataset enables for a more high level understanding of agent intent to mimic how they plan their trajectory. Figure~\ref{fig:accuracyhorizon} illustrates the performance of intention prediction over a 25 frame (5s) prediction horizon. Our work is the first to baseline both pedestrian and vehicle intent on a frame-wise level. We notice that prediction performance monotonically worsens over the horizon. However, we notice that for vehicles the intention accuracy in the multimodal setting are significantly improved from the unimodal case. This explains why intention conditioning helps more in the multimodal case, as agent intents are much more accurately understood. In contrast, only a slight improvement in intention performance for pedestrians. We posit this is because the intents for pedestrians do not change as frequently and are not as granular capturing direction such as "turn left"; thus, having more samples does not necessarily increase performance.

To better understand intention estimation, we visualize the confusion matrices as shown in Figure~\ref{fig:confusion_matrices}. For vehicles, we use the following set of discrete actions: \textit{moving},  \textit{stopped}, \textit{parking},  \textit{lane change},  \textit{turn left}, and  \textit{turn right}. We observed improved performance for vehicle intention prediction with multimodal goal destination sampling, indicating that our model can correlate long-term goals with short-term intent. For pedestrians, we use \textit{moving}, \textit{waiting to cross}, \textit{crossing}, and \textit{stopped}. The intents for pedestrians do not rapidly change unlike those for vehicles. Thus, we see that multimodal predictions do not actually improve pedestrian intention estimation. These results corroborate the results in Table \ref{tab:multimodaltable} where multimodal predictions with intention fail to outperform predictions without intentions. This is further examined in the next section.

\noindent \textbf{Effect of Annotation Frequency}:
Our dataset provides very detailed frame-wise intention labels at 5FPS for all agents. To examine the importance of having a dataset with such detailed annotations, we experiment with how changing annotation frequency can affect performance. We provide our model with oracle intentions available at varying frequencies. As shown in Figure ~\ref{fig:intfps}, trajectory prediction performance worsens roughly linearly as the frequency of intention labels reduces. This highlights the importance of our highly detailed annotations, as a choice to annotate every other frame (2.5FPS) clearly affects performance. Note that this effect is witnessed for primarily for vehicles, especially those that change lanes or turn. Pedestrian performance is not affected much, as the intention labels used for pedestrians do not change drastically for each frame. This also explains why intention conditioning did not help for multimodal evaluation for pedestrians as seen in Table \ref{tab:multimodaltable}.

\section{Conclusion}
In this work, we presented a large-scale heterogeneous dataset with detailed, frame-wise intention annotations. This dataset allows for both traditional trajectory prediction as well as understanding how intent changes over a long time horizon. In doing so, this dataset is this first that can be used as a benchmark for intention understanding for both vehicles and pedestrians. Furthermore, we formulate a joint trajectory and intention prediction framework which outperforms state-of-the-art on trajectory prediction metrics and offers a strong baseline for intention prediction. We bridge the gap between trajectory prediction and intention prediction and show that combining the two can better model agents' decision-making process, assisting in trajectory prediction. We believe our dataset can inspire future works that consider intention prediction in addition to traditional trajectory forecasting. Doing so can give more insight into models' decisions and will be critical in designing and maintaining a safe forecasting system.

\section*{Acknowledgement}
We thank our Honda Research Institute USA colleagues -- Behzad Dariush for his advice and support, Jiawei Huang for sensor calibration, and Huan Doung Nugen for data inspection and quality control. 

\clearpage

{\small
\bibliography{egbib}
}
\clearpage

\begin{strip}
\begin{center}
    \centering
    \resizebox{1\textwidth}{!}
  { \begin{tabular}{c|c|c|c}
Category &Set & Instances&Description \\
\hline\hline
\multirow{9}{*}{Intention Labels (Vehicle)}&Stopped&130743& \pbox{20cm}{The vehicle is stopped. This can happen in many scenarios such as stopping for \\
a traffic light, waiting to make a turn at an intersection, yielding for a pedestrian, etc.} \\ \cline{2-4}
&Parked&127150& \pbox{20cm}{The vehicle is parked along the street or parking lot} \\ \cline{2-4}
&Lane change to the left&2120& \pbox{20cm}{\multirow{2}{*}{The vehicle  is merging into the next lane.}} \\ \cline{2-3}
&Lane change to the right&2087& \\ \cline{2-4}
&Cut in to the left&347& \pbox{20cm}{\multirow{2}{*}{The vehicle is cutting into another lane.}} \\ \cline{2-3}
&Cut in to the right&736& \\ \cline{2-4}
&Turn left&15190& \pbox{20cm}{\multirow{2}{*}{The vehicle is turning (ex: at an intersection or towards a highway ramp).}} \\ \cline{2-3}
&Turn right&13171& \\ \cline{2-4}
&Moving / Other&206243& \pbox{20cm}{The vehicle is driving forward or some other movement that is not captured in the other labels.} \\ \hline
\multirow{5}{*}{Intention Labels (Pedestrian)}&Stopped&32538& \pbox{20cm}{The pedestrian is stopped along the street} \\ \cline{2-4}
&Moving&241889& \pbox{20cm}{The pedestrian is walking (ex: along the street)} \\ \cline{2-4}
&Waiting to cross&49576& \pbox{20cm}{The pedestrian is waiting to cross the intersection.} \\ \cline{2-4}
&Crossing the road&64870& \pbox{20cm}{The pedestrian is crossing the road.} \\ \cline{2-4}
&Potential Destination&67862& \pbox{20cm}{The potential location where the pedestrian may walk to.} \\ \hline
\multirow{5}{*}{Environmental Labels}&Lane information&440338&\pbox{20cm}{The possible actions a vehicle can take based on the current lane it 
is in. ($e.g.$ right turn, left\\ Turn, go forward, u-turn, lane change not possible). Note that multiple choices can be selected\\ depending on the situation. For example, a 
vehicle can be in a lane that goes forward or turns \\left. In our dataset, if a lane type is possible we select 1 and if it is not possible we select 0. So-\\metimes, if the vehicle
is out of frame and lane information cannot be deduced, we label it as -1.} \\ \cline{2-4}
&Traffic light&42476& \pbox{20cm}{The current state of the traffic light ($e.g.$ Red straight, Green round, Yellow round, etc.)} \\ \cline{2-4}
&Traffic sign&39066& \pbox{20cm}{The type of the traffic sign ($e.g.$ Stop, Left turn only, Do not enter for all)} \\ \cline{2-4}
&Road Exit and Entrance&126889& \pbox{20cm}{The positions of the road entrances/exits for a given scene. There can be a variable number\\ of road entrances/exits depending on map topology. Refer to figure \ref{fig:dataset_details} for more details.} \\ \hline
\multirow{4}{*}{Contextual Labels}&Age&166874& \pbox{20cm}{The estimated age category (child, adult, senior) of the pedestrian.} \\ \cline{2-4}
&Gender&166874& \pbox{20cm}{The gender of the pedestrian (male/female)} \\ \cline{2-4}
&Weather&644& \pbox{20cm}{The weather condition of the scenario (Sunny/Dusk/Cloudy/Night).} \\ \cline{2-4}
&Road condition&644& \pbox{20cm}{The road surface condition (dry / wet).} \\ \cline{2-4}
\hline
\end{tabular}}
    \captionof{table}{Details of the LOKI dataset. We report the various types of labels, number of instances of each label, and descriptions for all label types.}
    \label{tbl:datasetdetail}
\end{center}%
\end{strip}

\begin{figure*}[t!]
\begin{center}
     \centering
    \includegraphics[width=1.\textwidth]{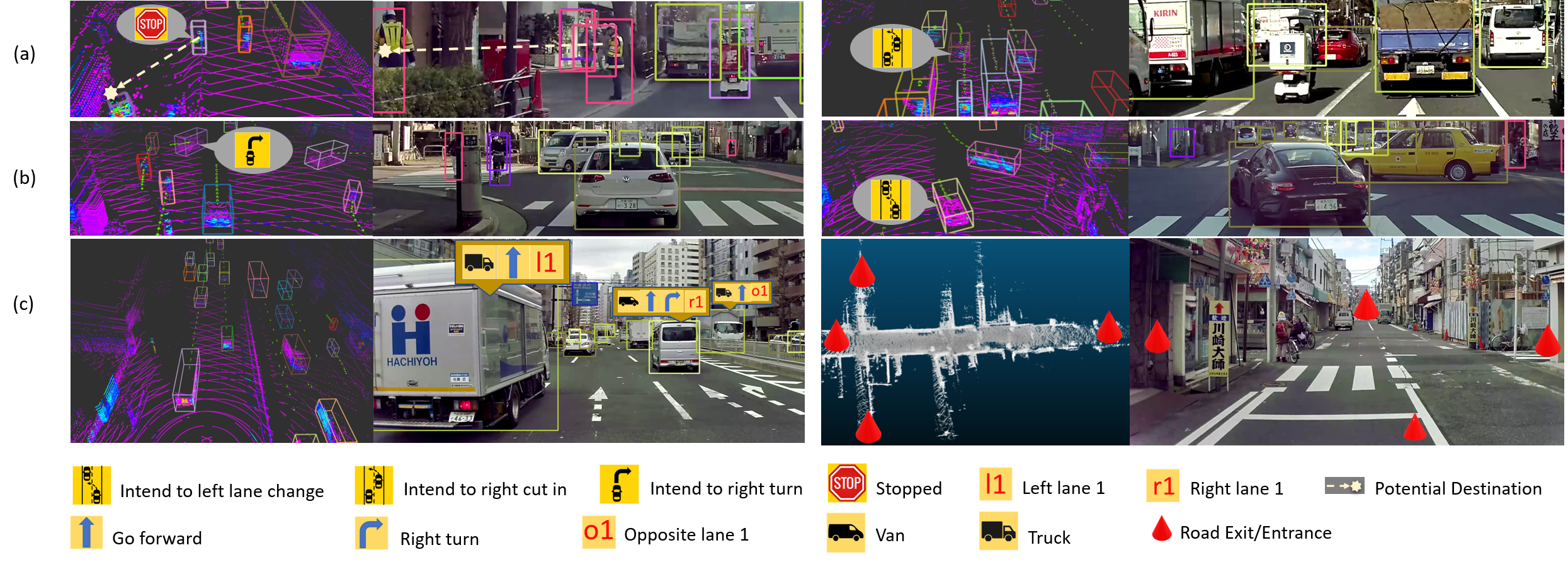}
    \caption{Visualization of three types of labels: (a-b) Intention labels; and (c) Environmental labels. The left part of each image is from laser scan and the right part is from RGB camera. In (a), the current status of pedestrian is "Stopped", and the potential destination shows where the pedestrian may go in the future. In (c) left, the blue arrow indicates the possible action of the vehicle based on the current lane it is on. The red words show the lane position related to the ego-vehicle.}
    \label{fig:dataset_details}
\end{center}%
\end{figure*}

\section{Details of the LOKI Dataset}

The LOKI dataset is collected from central Tokyo, Japan using an instrumented Honda SHUTTLE DBA-GK9 vehicle. Driving scenarios are collected from both suburban and urban areas at different times of the day. The camera, LiDAR, GPS and vehicle CAN BUS information were recorded. The RGB camera and four LiDAR sensors are placed on top of the vehicle to obtain better environment coverage. In addition, the timestamps were recorded for post multi-sensor synchronization processing. The cameras and LiDARs were placed on top of the vehicle to obtain better environment coverage. The sensors used for recording this dataset are listed below:

\begin{itemize}
\item A color SEKONIX SF332X-10X video camera (30HZ frame rate, $1928\times1280$ resolution and $60 \degree$ field-of-view (FOV)).

\item Four Velodyne VLP-32C 3D LiDARs (10 HZ  spin-rate, 32 laser beams, range: 200m, vertical FOV $40 \degree$).
\item A MTi-G-710-GNSS/INS-2A8G4 with output gyros, accelerometers and GPS.
\end{itemize} 

We used the CAN BUS to compensate for the ego motion while merging the LiDAR data and then transformed it to the virtual position (the center of the vehicle). The calibration is obtained through the extrinsic (the transformation between virtual LiDAR point and camera) and intrinsic camera parameters.

The recorded agents fall into 8 main classes. The vehicle classes are truck, van, car, bicyclist, motorcyclist, and bus. The pedestrian classes are pedestrian and wheelchair. As described in the main manuscript, we have three types of labels in the LOKI dataset: Intention labels, Environmental labels and Contextual labels. Intention labels for the vehicle classes include diverse motion state such as stop, lane change, cut in, etc., which can be observed in suburban and urban driving scenarios. Similarly, we annotated intention labels for the pedestrian classes such as moving, waiting to cross, etc. We additionally annotate a potential destination of stopped / waiting agents under the pedestrian classes, which is a direct indicative of their intention. Note that the potential destination cannot be obtained from the future location of agents as they mostly stay still until the end of the video clip. Details of the labels and their description are shown in Table~\ref{tbl:datasetdetail}. To further explore
how environments and scene context can affect the future behavior of agents, we annotate environmental labels (lane information, traffic light / sign, road entrance / exit) as well as contextual labels ( age, gender, weather, road condition). Figure~\ref{fig:dataset_details} illustrates different types of labels that we annotated in the LOKI dataset.

\section{Model Implementation}
In this section we provide more information on our data pre-processing, model choices and architecture details of each module.

\subsection{Data Pre-processing}
The LOKI dataset contains diverse traffic scenarios of up to 20 seconds, with the average recorded scene length of 12.6 seconds. With access to longer recordings, our dataset can be used for a multitude of trajectory prediction settings, from very short-term observations and predictions (3 seconds) to much longer observations and prediction horizons (10+ seconds). In our work we consider a long-term prediction setting with a $3s$ observation horizon and a $5s$ prediction horizon. Thus, we filter all agents that are not observed for at least $8s$ in a given traffic scenario. We use a sliding-window of $0.2s$ to augment tracklets. Furthermore, as in other works \cite{casas2018intentnet, salzmann2020trajectron++}, we further augment our dataset during training with ``rare" examples such as turning and lane changes. 

We solve the problem of intention prediction and trajectory prediction jointly as described in the main manuscript. The types of intentions for pedestrian agents and vehicle agents are different due to their different traffic restrictions and trajectory behaviors. For vehicles, we use the following set of discrete actions: \textit{Other/moving},  \textit{Stopped}, \textit{Parked},  \textit{Lane change},  \textit{Turn left}, and  \textit{Turn right}. We group \textit{Lane change to the left} and \textit{Lane change to the right} into a single intention type, as the number of instances that contain lane changes are much smaller and we noticed that separating the two did not improve performance. Furthermore, we do not compute a loss or predict on trajectories that contain \textit{Cut in to the left} and \textit{Cut in to the right}
, as we noticed that these constitute less than $0.01\%$ of the dataset, making it hard for the model to meaningfully distinguish from turning and lane-changing behavior. For pedestrians, we use the following set of intentions: \textit{Moving}, \textit{Waiting to cross}, \textit{Crossing the road}, and \textit{Stopped}. 

Our dataset originally contains frame-wise action labels for each agent. In order to use them as intention labels, we define intention to be a future action \cite{rasouli2021pedestrian}. Thus, the intention of an agent at frame $m$ is the agent's action at frame $m+q$ where we fix $q=4$ frames ($0.8s$) for our work. Note that for the observation period, we do not use intentions and only input observed actions to the model to prevent ground-truth leakage; the intention labels are only used for future trajectory prediction.

\subsection{Observation Encoder}
The observation encoder outputs a representation of past motion history, observed actions, and lane information for each actor independently. In our paper, we refer to past actions and lane information as observed states.

\begin{table}[h!]
\centering\scriptsize
\begin{tabular}{ c|l|c|c}
 &Layer & Input shape & Output shape\\
\hline\hline
0&encoder$\_$past.GRUCell.enc & [1, 15, 21] & [1, 64]\\
\hline
\end{tabular}
 \caption{We use a GRU to encode the observation information for each actor. We use a hidden dimension of 64. The input is 15 observation frames with 21 inputs at each frame (2 from position, 8 from vehicle actions, 5 from pedestrian actions, and 6 from lane information). We use one-hot-encoding to represent action types. We also include a "None" class for both vehicle and pedestrian actions. This allows vehicle agents to choose "None" for pedestrian action types and pedestrian agents to choose "None" for vehicle action types.}
  \label{tbl:fol_model}
\end{table}

\subsection{Long-term Goal Proposal Network}
For each actor, we first independently predict a proposed long-term goal position \cite{mangalam2020not, zhao2020tnt}. The proposed destination is similar as in other works  \cite{mangalam2020not} and is simply the endpoint of the trajectory, which in our case is $5s$ in the future. Because there are many plausible futures, we capture multimodality through learning a long-term goal distribution for each actor. To predict multiple trajectories, we sample various goals and condition our Scene Graph + Prediction decoder module on each sampled goal. We follow a similar formulation as proposed in \cite{mangalam2020not} and use a Conditional Variational Autoencoder (CVAE) to learn a latent distribution of the goals.

\begin{table}[h!]
\centering\scriptsize
\begin{tabular}{ c|l|c|c}
 &Layer & Input shape & Output shape\\
\hline\hline
0&encoder$\_$destination.Linear$\_$1 & [1, 2] & [1, 8]\\
1&encoder$\_$destination.ReLU$\_$1 & - &  -\\
2&encoder$\_$destination.Linear$\_$2 & [1, 8] & [1, 16]\\
3&encoder$\_$destination.ReLU$\_$2 & - &  -\\
4&encoder$\_$destination.Linear$\_$3 & [1, 16] & [1, 16]\\

5&encoder$\_$latent.Linear$\_$1 & [1, 80] & [1, 8]\\
6&encoder$\_$latent.ReLU$\_$1 & - &  -\\
7&encoder$\_$latent.Linear$\_$2 & [1, 8] & [1, 50]\\
8&encoder$\_$latent.ReLU$\_$2 & - &  -\\
9&encoder$\_$latent.Linear$\_$3 & [1, 50] & [1, 32]\\

10&decoder$\_$latent.Linear$\_$1 & [1, 80] & [1, 1024]\\
11&decoder$\_$latent.ReLU$\_$1 & - &  -\\
12&decoder$\_$latent.Linear$\_$2 & [1, 1024] & [1, 512]\\
13&decoder$\_$latent.ReLU$\_$2 & - &  -\\
14&decoder$\_$latent.Linear$\_$3 & [1, 512] & [1, 1024]\\
15&decoder$\_$latent.ReLU$\_$3 & - &  -\\
16&decoder$\_$latent.Linear$\_$4 & [1, 1024] & [1, 2]\\
\hline
\end{tabular}
 \caption{Sub-network architectures used for the goal-proposal network, modeled closely from model \cite{mangalam2020not}. Batch size of 1 used for example.}
  \label{tbl:longtermnetwork}
\end{table}

\subsection{Scene Graph + Prediction Module}
The Scene Graph and prediction module performs joint intention and trajectory prediction while reasoning about various factors that may affect agent intent including i) agent's own will ii) agent-agent interaction and iii) agent-environment influence. 

We construct a traffic scene graph to capture interaction and environmental influence. We have two types of nodes: 1) agent nodes 2) road entrance/exit nodes. The agent nodes are for dynamic agents in a scene (vehicles and pedestrians). The road entrance/exit nodes are static nodes that are positional markers that indicate where a road entrance or exit lies. These static nodes are used to provide information regarding map topology. In this work, we assume that these road markers are accessible to the model based on the annotations in our dataset. As described in our main manuscript, we use directional edges to propagate information through the various scene agents. Agents are connected to each other with bidirectional edges if they are within a certain threshold of $20$ meters away from each other. Similarly, we connect a directed edge from static nodes to dynamic nodes if the agent is within $35$ meters of the road entrance/exit. This graph is flexible in that a variable number of nodes or node types can be added as modification to this graph.

The Scene Graph + Prediction Module is then used to recurrently propagate information, predict intention, and predict trajectory. At each timestep, we first compute edge attributes between each pair of nodes. We use the edge$\_$attr network to embed nodes' velocities and relative positions between each pair of nodes. We then use the transformer$\_$conv layer (with a single attention head) \cite{transconv} for message passing and update each node's hidden states based on its neighbors. Following this, we use the vehicle$\_$intention$\_$predictor and pedestrian$\_$intention$\_$predictor networks to predict agent intention at that current timestep. The trajectory$\_$predictor is then conditioned on the hidden state of rnn$\_$future$\_$GRUCell.dec and the current intention prediction to predict the next position of each agent. Finally, the predicted positions are then inputted into rnn$\_$future$\_$GRUCell.dec to update the hidden states of each actor. This entire process is repeated for the prediction horizon length to unroll full trajectories while accounting for interactions and environmental information.

\begin{table}[h!]
\centering\scriptsize
\begin{tabular}{ c|l|c|c}
 &Layer & Input shape & Output shape\\
\hline\hline
0&trajectory$\_$predictor.Linear$\_$1 & [1, 93] & [1, 80]\\
1&trajectory$\_$predictor.ReLU$\_$1 & - &  -\\
2&trajectory$\_$predictor.Linear$\_$2 & [1, 80] & [1, 40]\\
3&trajectory$\_$predictor.ReLU$\_$2 & - &  -\\
4&trajectory$\_$predictor.Linear$\_$3 & [1, 40] & [1, 2]\\

5&vehicle$\_$intention$\_$predictor.Linear$\_$1 & [1, 80] & [1, 256]\\
6&vehicle$\_$intention$\_$predictor.ReLU$\_$1 & - & -\\
7&vehicle$\_$intention$\_$predictor.Linear$\_$2 & [1, 256] & [1, 128]\\
8&vehicle$\_$intention$\_$predictor.ReLU$\_$2 & - & -\\
9&vehicle$\_$intention$\_$predictor.Linear$\_$3 & [1, 128] & [1, 8]\\

10&pedestrian$\_$intention$\_$predictor.Linear$\_$1 & [1, 80] & [1, 256]\\
11&pedestrian$\_$intention$\_$predictor.ReLU$\_$1 & - & -\\
12&pedestrian$\_$intention$\_$predictor.Linear$\_$2 & [1, 256] & [1, 128]\\
13&pedestrian$\_$intention$\_$predictor.ReLU$\_$2 & - & -\\
14&pedestrian$\_$intention$\_$predictor.Linear$\_$3 & [1, 128] & [1, 5]\\

15&rnn$\_$future.GRUCell.dec & [1, 1, 80] & [1, 1, 80]\\

16&edge$\_$attr.Linear$\_$1 & [1, 8] & [1, 16]\\
17&edge$\_$attr.ReLU$\_$1 & - & -\\
18&edge$\_$attr.Linear$\_$2 & [1, 16] & [1, 16]\\

19&transformer$\_$conv$\_$1 & [1, 80] & [1, 80]\\

\hline
\end{tabular}
 \caption{Sub-network architectures used for the Scene Graph + Prediction module. Batch size of 1 used for example.}
  \label{tbl:longtermnetwork}
\end{table}

\subsection{Training}
\subsubsection{Loss Functions}
For convenience, we copy the loss functions used for training from our manuscript: 

\begin{equation*}
\mathcal{L}_{GPN} = \alpha_1
D_{KL}(\mathcal{N}({\mu}, {\sigma}) \Vert \mathcal{N}(0, {I})) + \alpha_2 \Vert\hat{G} - {G}\Vert^2_2
\end{equation*}

\begin{equation*}
\mathcal{L}_{int} = -\sum_{i=0}^{n} w_i*y_i*log(\hat{y_i})
\end{equation*}

\begin{equation*}
\mathcal{L}_{traj} = ||V-\hat{V}||_2
\end{equation*}

\begin{equation*}
\mathcal{L}_{Final} = \lambda_1 \mathcal{L}_{GPN} + \lambda_2\mathcal{L}_{int} + \lambda_3\mathcal{L}_{traj}
\end{equation*}

We set $\lambda_1=1$, $\lambda_2=100$, $\lambda_3=200$, $\alpha_1=1$, $\alpha_2=1$

\subsubsection{Training details}
We train the entire network end-to-end with the $\mathcal{L}_{Final}$ loss using a batch size of 32 scenarios and learning rate of $1\times10^{-4}$ using the ADAM optimizer. The intention prediction and trajectory forecasting tasks are heavily related with one another; thus we observed that training end-to-end helped with performance compared to modular training. Note that our batches are also grouped with an appropriate adjacency list to denote neighbors (connected edges) in a given batch. 

During training, we train with the ground-truth destination as the long-term goal, as we noticed that because short-term intentions are influenced by long-term goals, it is important for the intention prediction networks to get a clean signal while training. During testing, we condition on a sampled goal from the Goal-proposal Network. We also adopt the truncation trick as in \cite{mangalam2020not} to appropriately sample based on a varying number of future trajectories. The latent variable is sampled from different distributions depending on the number of future trajectories to be predicted: for $N=1$ (single-shot) we sample the from $\mathcal{N}(0,0)$ while for $N=20$ (multimodal) we sample from $\mathcal{N}(0,1.1)$.

\section{Visualizations}
\begin{figure}[!h]
    \centering
    \includegraphics[width=0.4\textwidth]{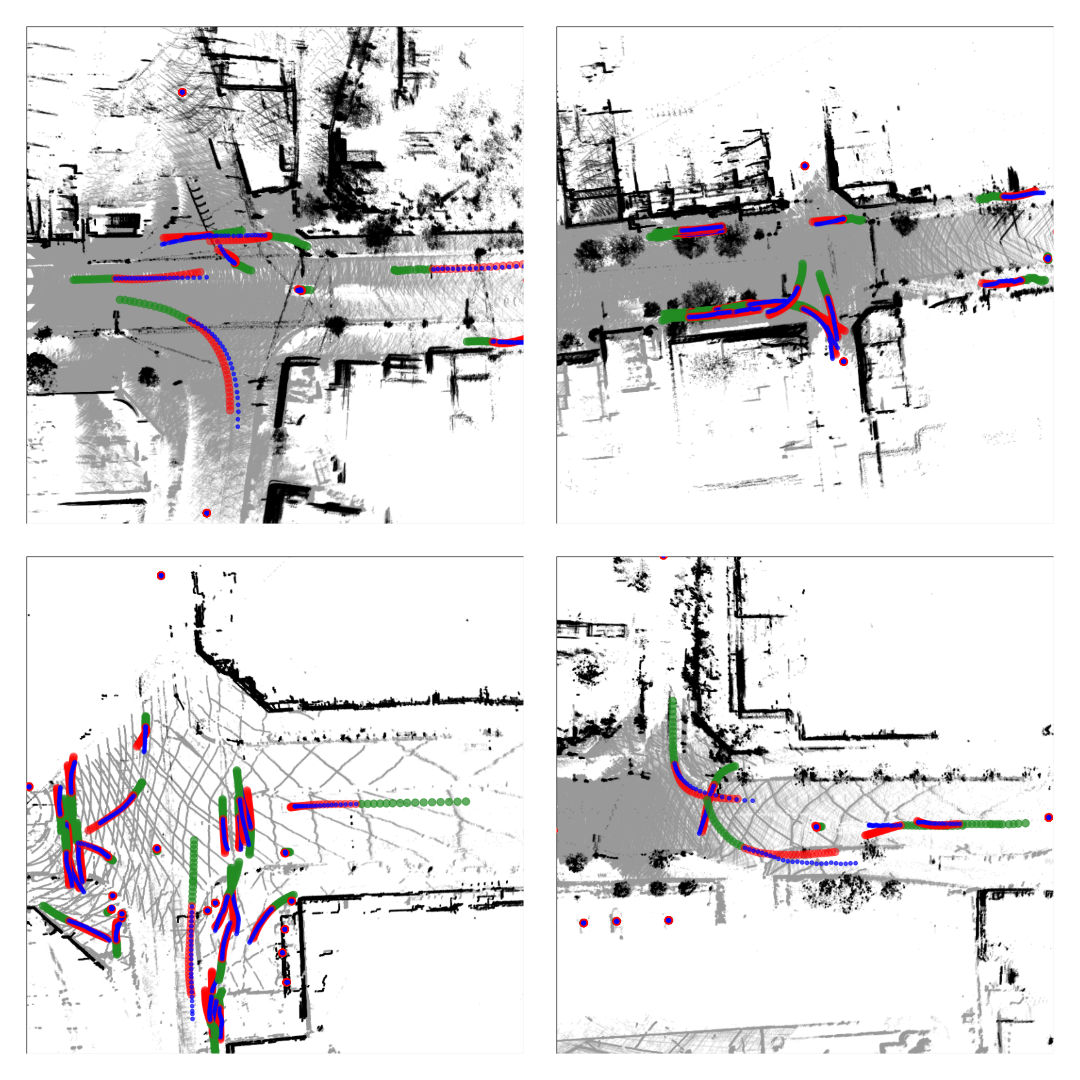}
    \caption{Visualization of our model's (Ours+IC+SG) top-1 (out of N=20 multimodal setting) predictions. Agent's past trajectory is represented in green. Agent's ground truth future is blue. Agent's predicted trajectories are in red (with increasing opacity to indicate better matches to the ground truth). We observe that our model performs reasonably in complex traffic scenarios.}
    \label{fig:top1}
\end{figure}

\begin{figure}[!h]
    \centering
    \includegraphics[width=0.41\textwidth]{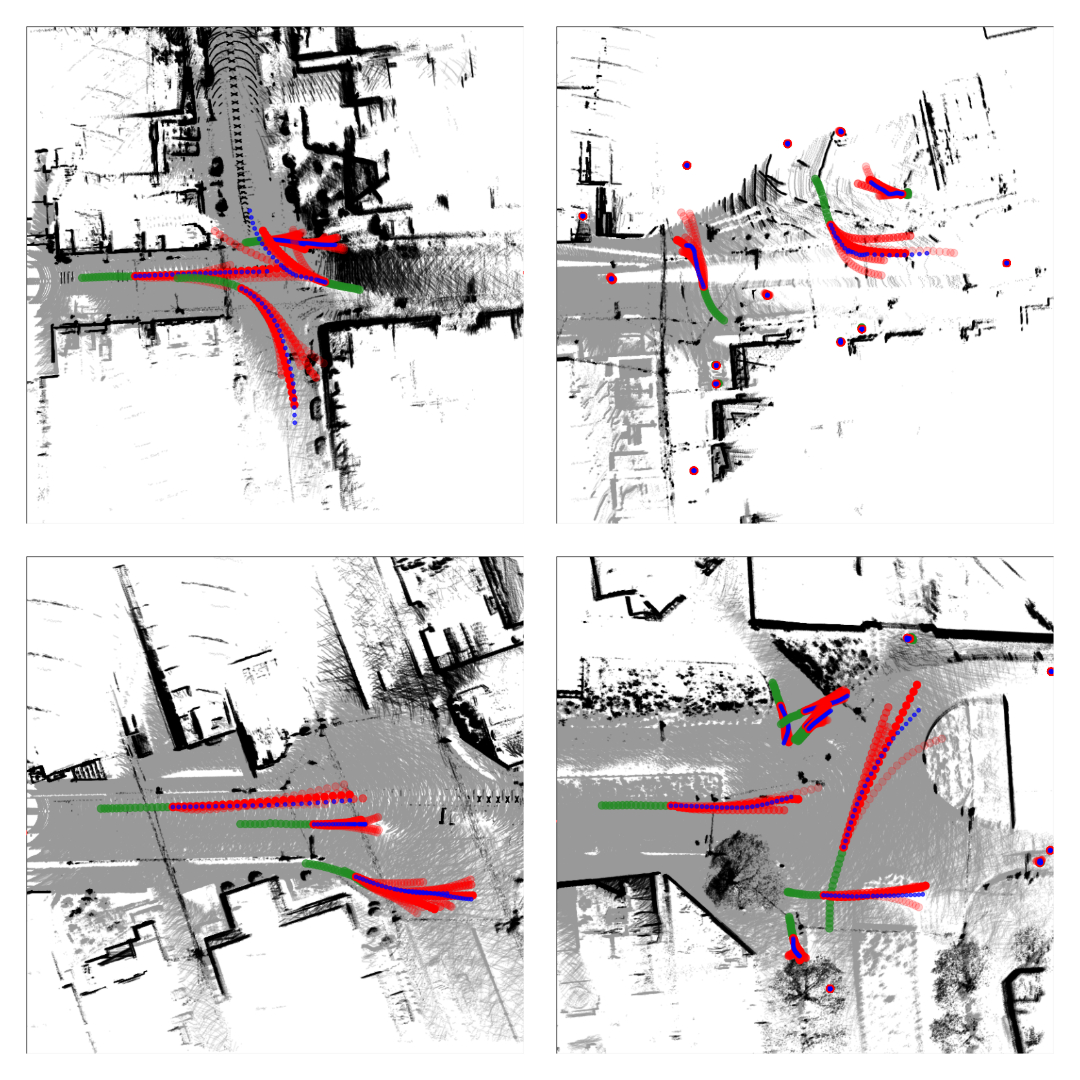}
    \caption{Visualization of our model's (Ours+IC+SG) top-5 (out of N=20 multimodal setting) predictions. Agent's past trajectory is represented in green. Agent's ground truth future is blue. Agent's predicted trajectories are in red (with increasing opacity to indicate better matches to the ground truth). }
    \label{fig:multimodal}
\end{figure}

In this section, we provide multiple visualizations that illustrate our proposed model's top-1 predictions (Figure~\ref{fig:top1}), top-5 multimodal predictions (Figure~\ref{fig:multimodal}), and our model's predictions with and without intention conditioning (Figure~\ref{fig:wo_intention}). Please view the video files provided in the supplementary folder for more detailed visualizations.

\begin{figure}[!h]
    \centering
    \includegraphics[width=0.45\textwidth]{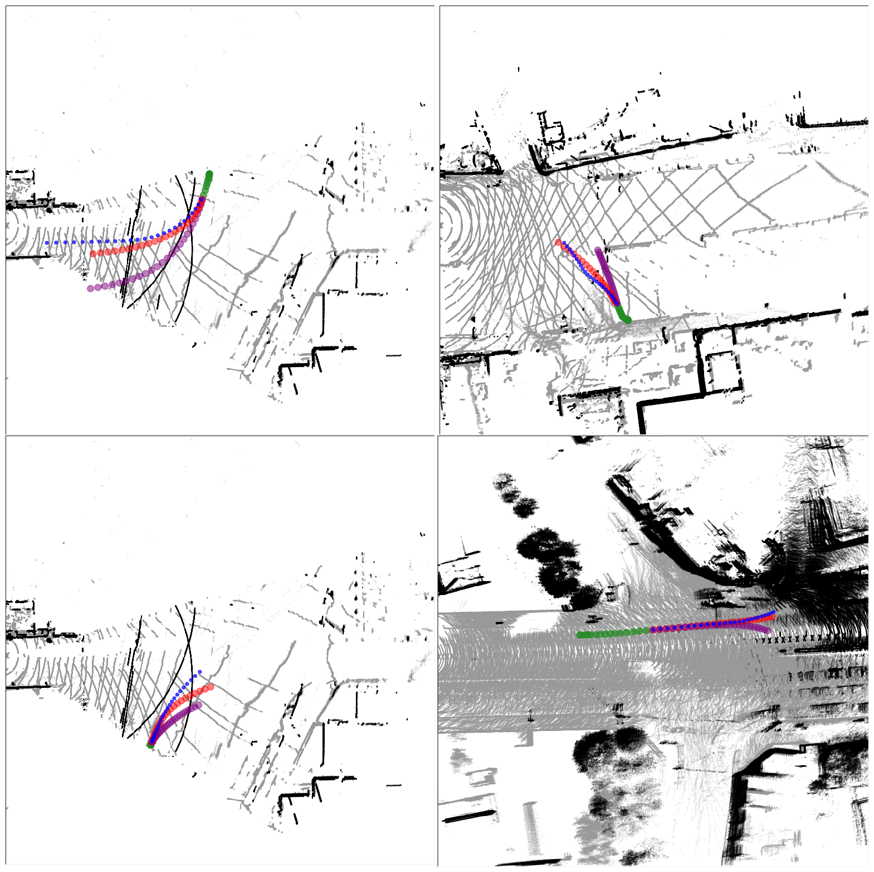}
    \caption{Comparison of with and without intention priors/scene graph for trajectory prediction. Agent's past trajectory is represented in green. Agent's ground truth future is blue. The top-1 predictions by the model without intention conditioning and scene graph are in purple. The top-1 predictions by the model with intention conditioning and scene graph are in red. We can qualitatively observe the efficacy of intention conditioning and incorporating interaction and environmental cues.}
    \label{fig:wo_intention}
\end{figure}

\end{document}